\documentclass[11pt]{article}

\usepackage[preprint]{acl}

\usepackage{times}
\usepackage{latexsym}

\usepackage[T1]{fontenc}

\usepackage[utf8]{inputenc}

\usepackage{microtype}

\usepackage{inconsolata}

\usepackage{graphicx}
\usepackage{xspace}

\newcommand{\fakeparagraph}[1]{\noindent\textbf{#1.}}
\newcommand{\modelname}{{DEER}\xspace}
\newcommand{\taskname}{{DG-MGT}\xspace}
\newcommand\figref[1]{Figure~\ref{#1}}

\newcommand\tabref[1]{Table~\ref{#1}}

\newcommand\secref[1]{Sec.~\ref{#1}}

\newcommand\appref[1]{Appendix~\ref{#1}}

\newcommand{\eg}{\emph{e.g.},\xspace}
\newcommand{\ie}{\emph{i.e.},\xspace}

\usepackage{xspace}
\usepackage{amsmath, amssymb, mathrsfs}
\usepackage{array}           
\usepackage{booktabs}        
\usepackage{colortbl}        
\usepackage[table]{xcolor}          
\usepackage{multirow}        
\usepackage{makecell}        
\usepackage{graphicx}        
\usepackage{rotating}        

\usepackage{algorithm}
\usepackage{algorithmic}
\usepackage[most]{tcolorbox}

\makeatletter
\newcommand{\Statex}{\State\hskip-\ALG@thistlm}
\makeatother

\newcommand{\Stage}[1]{%
  \vspace{0.4em}
  {\centering \textit{#1}\par}
  \vspace{0.2em}
}
\definecolor{Gray}{gray}{0.9}
\definecolor{RowGray}{gray}{0.9}
\definecolor{rowblue}{RGB}{228, 229, 252}

\title{DEER: Disentangled Mixture of Experts with Instance-Adaptive Routing for Generalizable Machine-Generated Text Detection}

\author{
    Guoxin Ma\textsuperscript{$1$},
    Xiaoming Liu\textsuperscript{$1$}\thanks{Corresponding author.},
    Hongyang Chen\textsuperscript{$2$},
    Chengzhengxu Li\textsuperscript{$1$} \\
    \textbf{Zhaohan Zhang\textsuperscript{$3$},
    Shengchao Liu\textsuperscript{$1$},
    Yu Lan\textsuperscript{$1$},
    Cong Wang\textsuperscript{$4$},
    Chao Shen\textsuperscript{$1$}} \\
    \textsuperscript{1}Faculty of Electronic and Information Engineering, Xi’an Jiaotong University\\ 
    \textsuperscript{2}China Mobile Group, 
    \textsuperscript{3}Queen Mary University of London, 
    \textsuperscript{4}City University of Hong Kong \\
    \texttt{\{guoxin.ma, czx.li, liusc\}@stu.xjtu.edu.cn} \\
    \texttt{zhaohan.zhang@qmul.ac.uk, \{xm.liu, ylan2020, chaoshen\}@xjtu.edu.cn}
}

\begin{document}
\maketitle
\begin{abstract}


Detecting machine-generated text has become a critical challenge amid the rapid advancement of LLMs, yet existing detectors degrade severely under domain shift. Through systematic pilot studies, we trace this vulnerability to two fundamental flaws in current generalization strategies, namely the incomplete preservation of domain-specific knowledge during multi-domain training and the misalignment between knowledge retrieval and the detection objective at inference.
To address these gaps, we propose \modelname{}, a \textbf{D}isentangled mixtur\textbf{E}-of-\textbf{E}xpe\textbf{R}ts framework that explicitly decouples domain-local and domain-invariant knowledge into specialized expert modules. 
Instead of static domain matching, \modelname{} employs a reinforcement learning-driven router that selects expert pathways based on instance-level detection rewards. 
This task-aligned, domain-agnostic mechanism ensures robust adaptation to unseen distributions by prioritizing detection utility over stylistic resemblance. 
Extensive experiments demonstrate that \modelname{} consistently outperforms state-of-the-art detectors, achieving average F1 improvements of \textbf{1.28\%} and \textbf{2.92\%}, and accuracy gains of \textbf{1.35\%} and \textbf{2.26\%} on in-domain and out-of-domain datasets,  offering reliable generalization for open-world deployment.

\end{abstract}

\section{Introduction}
In recent years, the rapid development and widespread adoption of large language models (LLMs), such as DeepSeek-R1\cite{guo2025deepseek}, GPT-5 \cite{openai2025gpt5}, and Claude-4.7-Opus \cite{Anthropic2026Claude}, have profoundly transformed the field of natural language generation. 
However, the unchecked use of LLMs has raised serious ethical and practical concerns, including the spread of misinformation and fake news \cite{liu2024does,liu2025mgt}, the generation of spam and malicious content \cite{ li2025ironsharpensirondefending, zellers2019defending}, and threats to academic integrity through AI-assisted ghostwriting \cite{verma2024ghostbuster}.
As a result, the development of accurate methods for detecting machine-generated text has emerged as a critical and urgent research priority.

Existing approaches often generalize poorly across domains because their learned representations are easily coupled with domain-specific biases \cite{liu-etal-2023-coco, li2025ironsharpensirondefending}. To improve generalizable detection, prior work mainly follows two multi-domain paradigms, \ie Domain-Invariant Generalization (DIG) and Domain-Matching Generalization (DMG). DIG trains a unified detector across multiple source domains and directly transfers the learned domain-invariant representations to unseen domains~\cite{bhattacharjee2024eagle, bhattacharjee2023conda, liu2025mgt}, while DMG models source-domain knowledge separately and retrieves relevant knowledge for an unseen-domain input through matching based on domain similarity~\cite{wu2025moses, wong2026k}. Despite their different strategies, both DIG and DMG rely on multi-domain training to improve unseen-domain detection. 
Motivated by this observation, we investigate a critical question for generalizable MGT detection: \textit{``Can existing paradigms effectively learn and utilize multi-domain knowledge under domain shift?''} Our pilot study in \secref{sec:pilot} reveals two key findings: 

\begin{tcolorbox}[
    colback=gray!5, 
    colframe=gray!20, 
    arc=3pt, 
    left=5pt, right=5pt, top=4pt, bottom=4pt
]
\noindent\textbf{$\mathcal{F}_1$:} 
Unified multi-domain training inherently dilutes domain-specific detection cues.
\vspace{4pt}

\noindent\textbf{$\mathcal{F}_2$:} 
Similarity-based routing misaligns knowledge utilization with the detection objective.
\end{tcolorbox}

These findings motivate us to propose \modelname{}, a two-stage framework for generalizable MGT detection. 
To address the knowledge dilution identified in $\mathcal{F}_1$, \modelname{} first introduces a \textbf{D}isentangled \textbf{M}ixture-\textbf{o}f-\textbf{E}xperts (DMoE) architecture that structurally decouples domain-specific experts (trained on individual source domains) from domain-shared experts (trained across all domains). 
This design preserves the unified modeling principle of DIG while explicitly retaining fine-grained domain-local cues that are typically suppressed. 
To resolve the routing misalignment highlighted in $\mathcal{F}_2$, \modelname{} second replaces the similarity-matching strategy of DMG with a reinforcement learning-driven routing mechanism. 
By optimizing expert selection directly from instance-level detection rewards, the router dynamically coordinates pathways that maximize task utility rather than stylistic resemblance. 
Extensive experiments across five in-domain (IND) and five out-of-distribution (OOD) benchmarks demonstrate that \modelname{} consistently outperforms state-of-the-art detectors, achieving average F1 improvements of \textbf{1.28\%} (IND) and \textbf{2.92\%} (OOD), alongside accuracy gains of \textbf{1.35\%} and \textbf{2.26\%}, respectively.
Our contributions are summarized as follows:

\begin{itemize}
    \item \textbf{Empirical Insight into DIG-DMG Limitations:} 
    We systematically expose two root causes of poor generalization in MGT detection, namely gradient-induced knowledge dilution in unified training (DIG) and task-agnostic routing misalignment in similarity-based matching (DMG).
    
    \item \textbf{Generalizable Detection Framework:} 
    We propose \modelname{}, which structurally decouples domain-local and domain-invariant knowledge via a Disentangled MoE architecture, and replaces static similarity matching with an RL-driven router that optimizes expert selection directly from instance-level detection rewards.
    
    \item \textbf{Outstanding Task Performance:} 
    \modelname{} achieves SOTA performance across diverse OOD benchmarks with significantly lower variance. Ablation studies confirm component efficacy, while incremental domain adaptation experiments demonstrate that the modular design enables efficient and stable adaptation to new domains.
\end{itemize}

\section{Related Work}

\fakeparagraph{DG-MGT Detection} 
Reliable MGT detection on unseen domains relies on metric-based or fine-tuned approaches. Metric methods \cite{gehrmann2019gltr,guo2024biscope,bao2024fastdetectgpt,wang2023seqxgpt,hans2024spotting} offer transferability but suffer threshold miscalibration under stylistic shifts; even conditional thresholding \cite{wu2025moses} only adjusts decisions post-hoc without resolving feature entanglement. Fine-tuned methods \cite{liu-etal-2023-coco,liu2024does,li2025ironsharpensirondefending,shum2023automatic,bhattacharjee2023conda,tan2022domain,liu2025mgt} learn domain-invariant features via adversarial or contrastive objectives. However, unified multi-domain training underutilizes domain-specific cues and ignores cross-domain heterogeneity that causes optimization interference. We address this via a disentangled framework that separately models domain-shared signatures and domain-local characteristics.

\fakeparagraph{Mixture of Experts} 
MoE enhances capacity through conditional computation \cite{jacobs1991adaptive} and has been adapted for domain generalization via expert specialization and aggregation~\cite{chen2024lfme,xu2024cbdmoe,zhong2022meta,qu2022hmoe,radwan2025feddg}. Yet, existing routing relies on coarse domain signals or static similarity~\cite{dai2021generalizable,qu2022hmoe,radwan2025feddg,ren2023pangu,wen2025measure}, limiting input-level adaptability. In contrast, we propose an instance-adaptive routing mechanism that dynamically selects experts based on context-aware detection utility, ensuring robust generalization under distribution shift.

\section{Pilot Study}\label{sec:pilot}
We conducted pilot studies to examine how existing paradigms learn and utilize multi-domain knowledge under distribution shift.
Specifically, we investigate whether unified training truly subsumes single-domain expertise (Sec.~\ref{sec:pilot_dmoe}) and whether domain similarity reliably guides knowledge selection for unseen inputs (Sec.~\ref{sec:pilot_rl}). These empirical observations directly motivate the architectural and routing designs of our framework.

\subsection{Knowledge Dilution in Unified Training}\label{sec:pilot_dmoe}

\fakeparagraph{Setup} To examine whether unified multi-domain training (DIG) fully preserves the discriminative knowledge captured by single-domain specialists, we construct a controlled comparison between a multi-domain (MD) detector and domain-specific detectors. Specifically, the MD detector is trained on mixed data from two source domains, while two specialist detectors are trained separately on the corresponding individual domains, using the same backbone and training protocol. All detectors are evaluated on held-out test sets from the seen source domains (IND) and the same unseen target domain (OOD), allowing us to compare their detection coverage under both settings.

\begin{figure}[h]
  \centering
  \includegraphics[width=\linewidth]{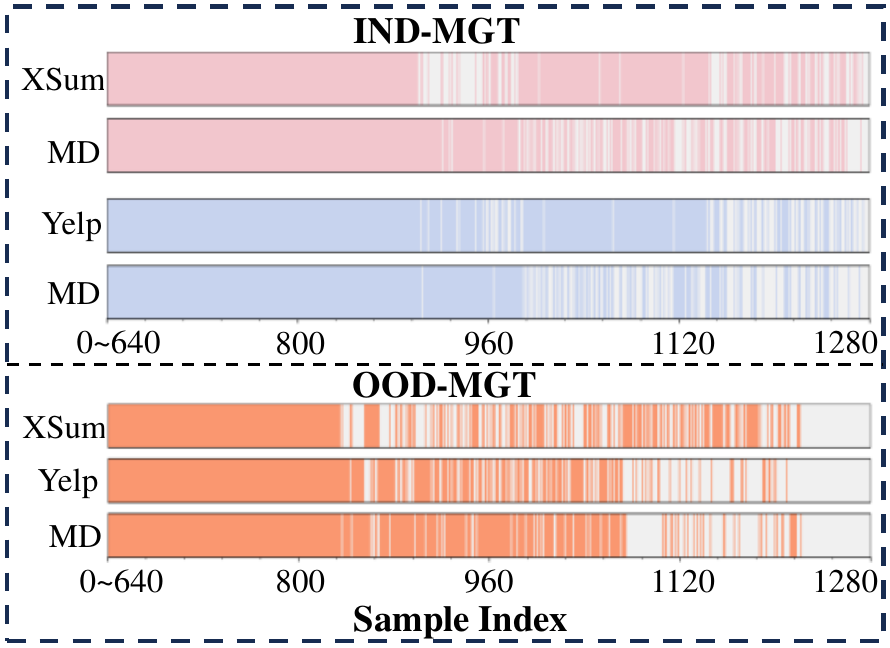}
  \caption{
    \textbf{Detection coverage under IND and OOD settings.}
    \textbf{Top:} evaluation on seen source-domain test sets.
    \textbf{Bottom:} evaluation on the same unseen target-domain test set.
    Colored and gray segments denote correctly detected and missed samples, respectively. Bars with the same color correspond to the same test set.
    }
  \label{fig:detect_coverage} 
  \vspace{-7pt}
\end{figure}

\fakeparagraph{Analysis} As shown in \figref{fig:detect_coverage}, the detection coverage analysis reveals two critical patterns. \textit{First}, despite using more training data, the MD detector is hindered by cross-domain heterogeneity in the IND setting, allowing single-domain specialists to achieve better coverage on their matched test sets. \textit{Second}, under OOD shifts, while the MD detector captures strong transferable signals, it still leaves a non-trivial fraction of instances undetected that are only recoverable by domain-specific detectors. 

\fakeparagraph{Implication} 
These results indicate that joint multi-domain optimization can dilute domain-local decision boundaries, leaving complementary cues underused for OOD detection.
Effective generalization therefore requires an architecture that explicitly decouples and preserves both domain-invariant signatures and domain-local distinctions, rather than forcing them into a single shared representation. This directly motivates our disentangled expert design in Sec.~\ref{sec:DMoE}.

\subsection{Routing Misalignment in Similarity Matching}\label{sec:pilot_rl}

\fakeparagraph{Setup} We next investigate whether domain similarity reliably identifies the most useful source knowledge for unseen inputs, as assumed by DMG paradigms. We train five single-domain detectors, each on one source domain, using the same backbone and training protocol. By applying all source-domain detectors to each OOD test sample, we partition samples into all, some, and none groups according to their detection outcomes. We then focus on the ``some'' group, where source-domain selection is necessary, and test whether routing to the most stylistically similar source domain yields the correct detection outcome.

\begin{figure}[h]
  \centering
  \includegraphics[width=1\linewidth]{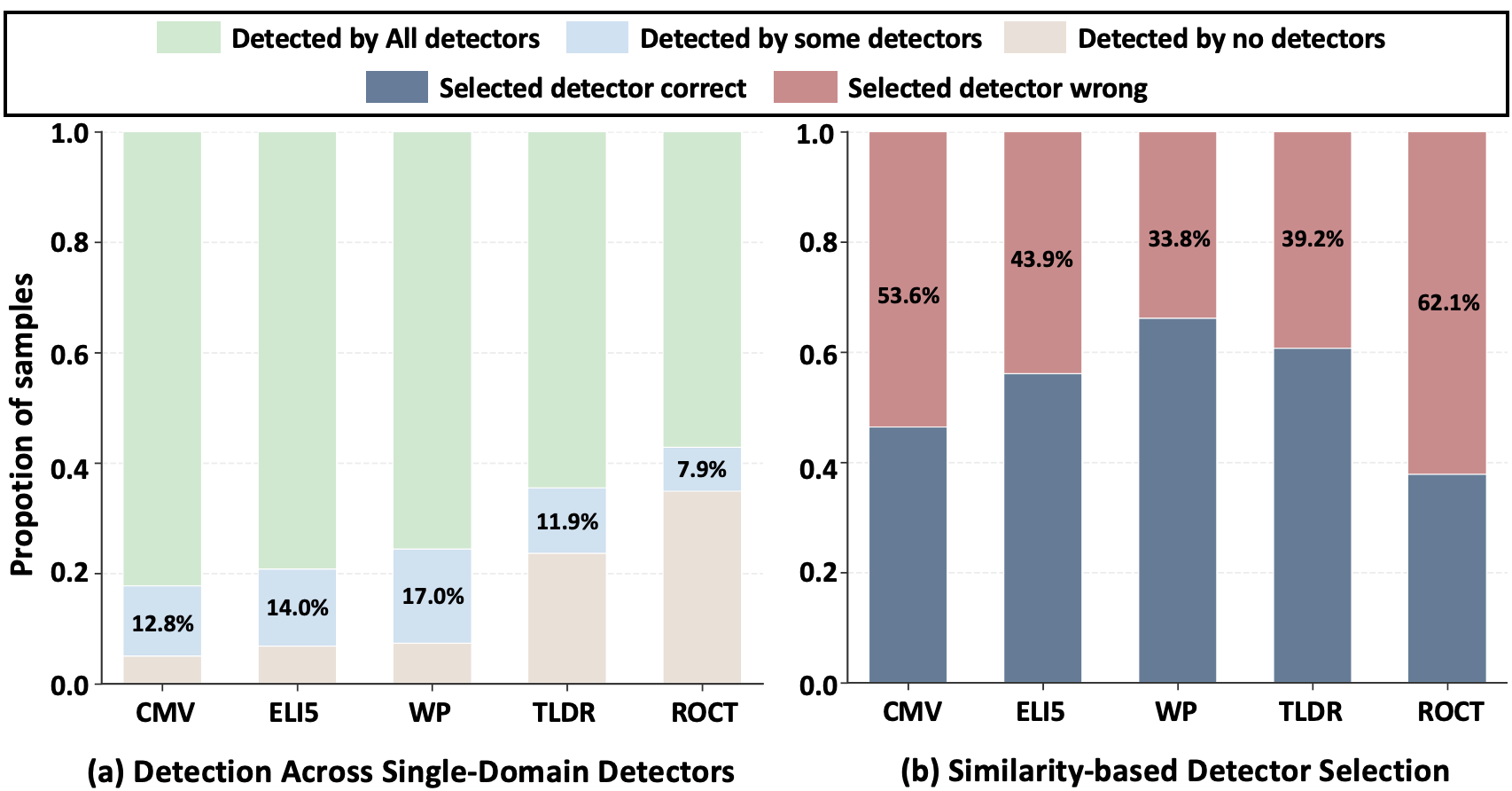}
  \caption{
\textbf{Instance-level knowledge utilization under domain shift.}
\textbf{(a)} OOD samples are divided into three groups according to whether they are correctly detected by all, some, or none of the source-domain detectors.
\textbf{(b)} Whether the most similar source-domain detector correctly predicts samples in the ``some'' group.
  }
  \label{fig:pilot_rl} 
  \vspace{-10pt}
\end{figure}

\fakeparagraph{Analysis} The selection analysis in \figref{fig:pilot_rl} yields two key observations. \textit{First}, a non-negligible portion of OOD samples is correctly classified by only a subset of source-domain detectors, indicating that detection utility can vary across source domains at the instance level. \textit{Second}, for these ambiguous instances, selecting the detector from the most similar source domain \textit{fails} to guarantee correct predictions. This demonstrates that surface-level domain relevance is a poor proxy for actual detection utility~\cite{zhang2021quantifying}.

\fakeparagraph{Implication} Similarity-based routing inherently misaligns knowledge selection with the downstream detection objective. Under distribution shift, static matching prioritizes stylistic resemblance over task utility, leading to suboptimal expert activation. To reliably orchestrate multi-domain knowledge, routing must be optimized directly against detection rewards rather than domain proximity, motivating our reinforcement learning-driven routing mechanism in Sec.~\ref{sec:RL_route}.

\begin{figure*}[ht]
  \centering
  \includegraphics[width=\textwidth]{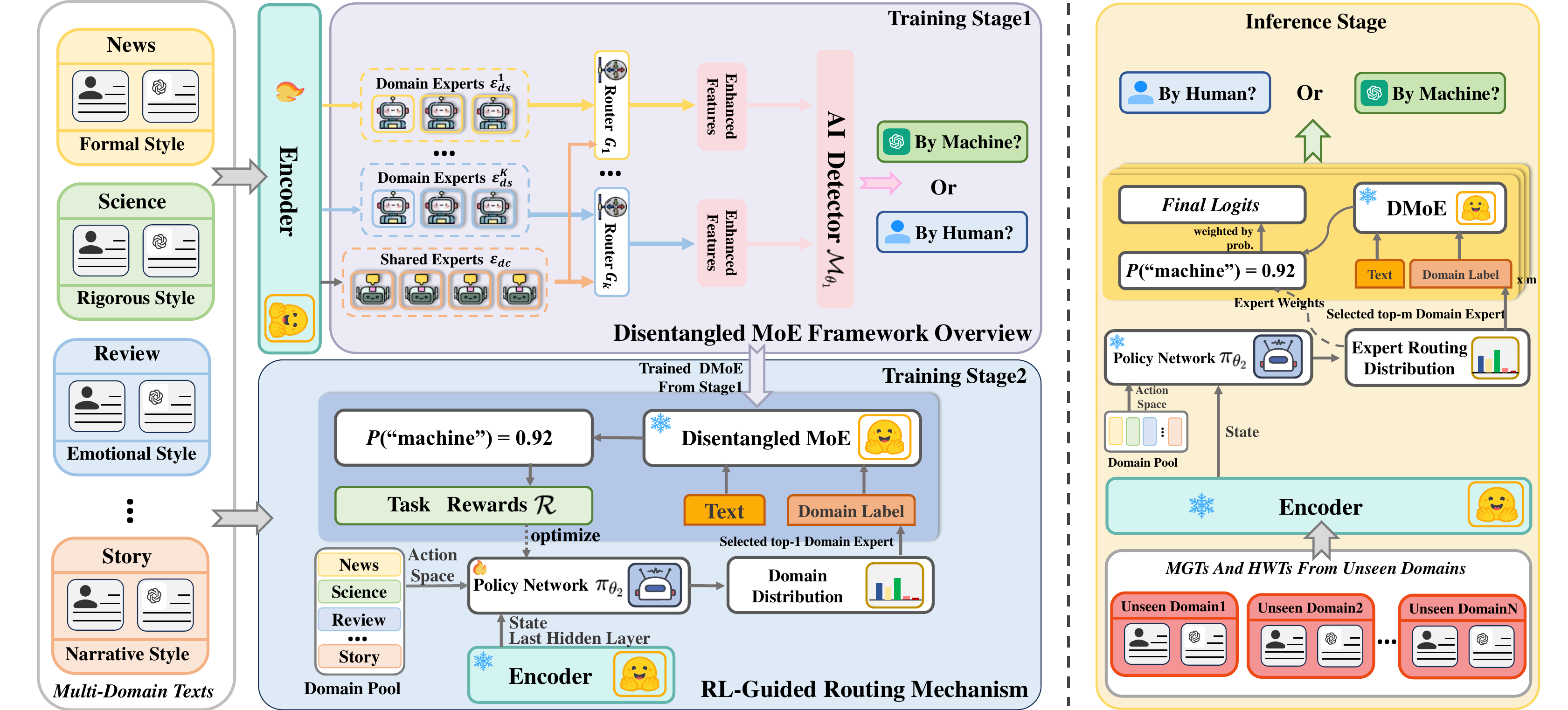} 
  \caption{\textbf{Overview of \modelname{}.} 
  \textbf{Left: Two-stage training.} Upper: DMoE learns domain-local patterns via domain-specific experts and transferable signatures via shared experts under domain supervision (\secref{sec:DMoE}). Lower: Frozen DMoE guides an RL policy to learn domain-agnostic routing from detection rewards (\secref{sec:RL_route}). 
  \textbf{Right: Inference.} The policy selects top-$m$ high-utility pathways; their outputs are adaptively fused with shared experts to produce the final prediction, enabling instance-aware generalization to unseen domains.}
  \label{fig:overall}
  \vspace{-10pt}
\end{figure*}

\section{Methodology}



Guided by the empirical observations in \secref{sec:pilot}, we propose \modelname{}, a two-stage framework designed for generalizable machine-generated text detection under domain shift.
The overall framework of \modelname{} is illustrated in Figure~\ref{fig:overall}.



\subsection{Disentangled Mixture-of-Experts Framework}\label{sec:DMoE}
Motivated by the observation in \secref{sec:pilot_dmoe} that unified training inherently dilutes domain-specific cues, we propose the \textbf{D}isentangled \textbf{M}ixture-\textbf{o}f-\textbf{E}xperts (DMoE) framework. 
To prevent the loss of fine-grained discriminative signals under joint optimization, DMoE structurally decouples domain-specific and domain-shared knowledge into isolated learning pathways. 
The architecture primarily comprises  \textit{expert modules} for specialized representation learning and a \textit{domain-aware gating function} for adaptive knowledge aggregation.

\fakeparagraph{Expert Modules}
Unlike conventional Multi-gate Mixture-of-Experts (MMoE) frameworks~\cite{ma2018modeling} that process training data uniformly, DMoE explicitly partitions data flow by domain to isolate and preserve both domain-specific and domain-invariant knowledge. Formally, given $n$ source domains $\mathcal{D}_{\text{train}} = \{ \mathcal{D}^k \}_{k=1}^n$, we instantiate a dedicated set of \textit{domain-specific experts} $\{ \mathcal{E}_{\text{ds}}^k \}_{k=1}^n$. Each $\mathcal{E}_{\text{ds}}^k$ comprises $m_1$ independently parameterized MLPs $\{ e_{\text{ds}}^{k,i} \}_{i=1}^{m_1}$ tasked with capturing fine-grained, domain-local discriminative cues. Crucially, $\mathcal{E}_{\text{ds}}^k$ is updated exclusively on samples from $\mathcal{D}^k$, preventing cross-domain gradient interference and directly mitigating the knowledge dilution identified in \secref{sec:pilot_dmoe}.

Complementing these, we introduce a \textit{domain-shared expert} $\mathcal{E}_{\text{dc}} = \{ e_{\text{dc}}^j \}_{j=1}^{m_2}$ to extract transferable, domain-invariant machinization signatures. Structurally analogous to the domain-specific modules, $\mathcal{E}_{\text{dc}}$ is trained on the union of all source domains. This design explicitly preserves the unified modeling principle of DIG while avoiding the systematic suppression of local decision boundaries.

\fakeparagraph{Domain-Aware Gating Function}
To address the knowledge dilution identified, DMoE employs a domain-aware gating mechanism that dynamically balances domain-local and domain-invariant cues. 
Given a training sample $(x_k, y_k, d_k) \in \mathcal{D}_{\text{train}}^k$ from the $k$-th source domain, the input text $x_k$ is first encoded into a contextual embedding $\mathbf{h}_k$ via a pre-trained encoder (\eg RoBERTa~\cite{liu2019roberta}):
\begin{equation}\label{eq:encoding}
    \mathbf{h}_k = \text{Encoder}(x_k).
\end{equation}

Unlike conventional MoE routers that rely on input similarity, our gating network $G_k(\cdot)$ is explicitly conditioned on the source domain. It projects $\mathbf{h}_k$ into a logit space via a learnable matrix $\mathbf{U}_k$, followed by Softmax normalization to yield a routing distribution $\mathbf{w}_k$ over the $m_1$ domain-specific and $m_2$ shared experts:
\begin{equation}\label{eq:gating_mechanism}
\begin{split}
    \mathbf{w}_k = G_k(\mathbf{h}_k) = \text{Softmax}(\mathbf{U}_k \mathbf{h}_k) \\
    = [\underbrace{w_k^{1}, \dots, w_k^{m_1}}_{\text{domain-specific}}, \underbrace{w_k^{m_1+1}, \dots, w_k^{m_1+m_2}}_{\text{domain-shared}}].
\end{split}
\end{equation}
The final representation $\mathbf{H}_k$ is then computed as the weighted aggregation of expert outputs:
\begingroup
\small
\begin{equation}\label{eq:aggregation}
    \mathbf{H}_k =
    \sum_{i=1}^{m_1} w_k^{i} \cdot \mathcal{E}_{\text{ds}}^{k,i}(\mathbf{h}_k)
    +
    \sum_{j=1}^{m_2} w_k^{m_1+j} \cdot \mathcal{E}_{\text{dc}}^{j}(\mathbf{h}_k).
\end{equation}
\endgroup

This fused representation is passed to a lightweight classification head $\mathcal{M}_{\theta_{1}}(\cdot)$ to predict the label $\hat{y}_k = \mathcal{M}_{\theta_{1}}(\mathbf{H}_k)$, where $\theta_{1}$ denotes the classifier parameters. The detection loss (\eg cross-entropy) between $\hat{y}_k$ and the ground truth $y_k$ drives gradient backpropagation to jointly update the encoder, all expert modules, and the gating networks.

By routing samples from domain $k$ through its domain-aware gate $G_k(\cdot)$, DMoE separates domain-local and shared optimization signals. Domain-specific experts learn discriminative cues only from their matched domains, while shared experts capture transferable patterns across domains. This design mitigates the knowledge dilution observed in Sec.~3.1 and provides complementary expert knowledge for OOD detection.



\subsection{RL-Guided Expert Routing Mechanism}\label{sec:RL_route}
Following expert specialization, we introduce a reinforcement learning-driven routing mechanism to dynamically coordinate expert pathways for unseen-domain inputs. 
Addressing the routing misalignment identified in \secref{sec:pilot_rl}, this component replaces static similarity matching~\cite{dai2021generalizable,wen2025measure} with a policy that optimizes expert selection directly against the detection objective. 
Formulated as an instance-level decision process, it trains a domain-agnostic policy network to activate pathways based on task-aligned utility rather than stylistic proximity.

\fakeparagraph{Policy Optimization}
Guided by the observation in \secref{sec:pilot_rl} that domain relevance poorly proxies detection utility, we formulate expert routing as an instance-level decision process. For an input $x_t$, the state $\mathbf{s}_t$ is defined as the contextual embedding extracted by the frozen DMoE encoder (Eq.~\ref{eq:encoding}). Conditioned on $\mathbf{s}_t$, the policy network $\pi_{\theta_2}(\cdot \mid \mathbf{s}_t)$ parameterizes a categorical distribution over the action space $\mathcal{A}=\{1,\dots,n\}$, where selecting action $a_t=k$ activates the $k$-th domain-conditioned pathway. The routing distribution is computed via a two-layer MLP:
\begin{equation}
\pi_{\theta_2}(a_t \mid \mathbf{s}_t) = \mathrm{softmax}\!\left( \mathbf{W}_2 \tanh(\mathbf{W}_1 \mathbf{s}_t) \right), \label{eq:policy}
\end{equation}
where $\mathbf{W}_1$ and $\mathbf{W}_2$ are learnable weight matrices. Upon sampling an action, the policy forwards $x_t$ through the corresponding DMoE pathway to obtain task-specific logits.

To align routing with the downstream objective, we optimize $\pi_{\theta_2}$ via policy gradient algorithm~\cite{sutton1988learning} using a margin-based reward signal. The immediate reward $r_t$ quantifies the detection confidence margin of the activated pathway:
\begin{equation}
r_t = p_{y_t}^{a_t} - p_{\bar{y}_t}^{a_t}, \label{eq:reward}
\end{equation}
where $y_t$ is the ground-truth label, $\bar{y}_t$ denotes its inverse, and $p_{y_t}^{a_t}$ is the probability assigned to $y_t$ under pathway $a_t$. This formulation explicitly incentivizes pathways that maximize classification certainty.

Because pathway utility varies significantly across instances, we evaluate the selected action relative to all candidate pathways. Specifically, we perform a full rollout over $\mathcal{A}$ and compute an instance-specific baseline $\bar{r}_t$ as the average reward:
\begin{equation}
\bar{r}_t = \frac{1}{n}\sum_{k=1}^{n} r_t^{k}, \label{eq:baseline}
\end{equation}
where $r_t^{k}$ is the reward obtained by activating the $k$-th pathway. The final advantage signal is defined as the relative reward:
\begin{equation}
r_{\text{final}}(x_t, a_t) = r_t - \bar{r}_t. \label{eq:rfinal}
\end{equation}
This baseline-subtracted signal stabilizes gradient updates and encourages the router to favor pathways whose detection rewards exceed the instance-level average, thereby aligning routing decisions with the detection objective. Implementation details and the complete training procedure are provided in \appref{A.3}, and \ref{A.4}.

\begin{table*}[ht]
\small
\centering
\resizebox{\textwidth}{!}{
\setlength{\tabcolsep}{3pt}
\begin{tabular}{l c cccccc cccccc}
\toprule
\multirow{2}{*}{\textbf{Methods}} 
& \multirow{2}{*}{\textbf{Metric}} 
& \multicolumn{6}{c}{\textbf{IND-MGT}} 
& \multicolumn{6}{c}{\textbf{DG-MGT}} \\
\cmidrule(lr){3-8} \cmidrule(lr){9-14}
& 
& \textit{\textbf{XSum}} & \textit{\textbf{HSwag.}} & \textit{\textbf{SQuAD}} & \textit{\textbf{Yelp}} & \textit{\textbf{Sci}} & \textit{\textbf{Avg.}}
& \textit{\textbf{CMV}} & \textit{\textbf{ELI5}} & \textit{\textbf{WP}} & \textit{\textbf{TLDR}} & \textit{\textbf{ROCT}} & \textit{\textbf{Avg.}} \\
\midrule
\rowcolor{rowblue}
\multicolumn{14}{c}{\textbf{\textit{Metric-based methods}}} \\
\midrule
\multirow{2}{*}{\begin{tabular}[c]{@{}l@{}}Fast-\\DetectGPT\end{tabular}}
& Acc 
& 51.17 & 78.79 & 75.65 & 71.76 & 73.91 & 70.26 
& 73.27 & 75.56 & 62.62 & 71.44 & \textbf{66.56} & 69.89 \\
& F1  
& 54.01  & 77.43  & 76.27  & 71.24  & 74.77  & 70.74 
& 75.00  & 73.60  & 62.15  & 71.73  & 70.40  & 70.58 \\
\midrule
\multirow{2}{*}{Binoculars} 
& Acc 
& 71.77 & 72.06 & 79.65 & 76.89 & 83.75 & 76.82 
& 86.92 & 83.99 & 82.83 & 72.31 & 60.50 & 77.31 \\
& F1  
& 66.77  & 73.84  & 79.29  & 76.22  & 82.43  & 75.71 
& 86.61  & 83.18  & 81.68  & 73.25  & 68.17  & 78.58 \\
\midrule
\rowcolor{rowblue}
\multicolumn{14}{c}{\textbf{\textit{Model-based methods}}} \\
\midrule

\multirow{2}{*}{Roberta$^\dag$} 
& Acc 
& 88.77$_{\text{0.64}}$ & 95.24$_{\text{1.07}}$ & 92.40$_{\text{0.66}}$ & 89.31$_{\text{3.09}}$ & 91.53$_{\text{1.78}}$ & 91.45
& 79.32$_{\text{3.96}}$ & 83.42$_{\text{1.80}}$ & 80.64$_{\text{3.39}}$ & 69.05$_{\text{1.89}}$ & 61.08$_{\text{4.03}}$ & 74.70 \\
& F1  
& 89.61$_{\text{0.48}}$ & 95.35$_{\text{1.18}}$ & 92.74$_{\text{0.54}}$ & 90.05$_{\text{2.46}}$ & 91.93$_{\text{1.48}}$ & 91.94 
& 83.12$_{\text{2.70}}$ & 85.45$_{\text{1.35}}$ & 83.68$_{\text{2.39}}$ & 75.88$_{\text{1.10}}$ & 71.86$_{\text{2.10}}$ & 79.99 \\
\midrule
\multirow{2}{*}{Ghostbuster} 
& Acc 
& 82.44$_{\text{7.94}}$ & 90.79$_{\text{5.58}}$ & 89.99$_{\text{3.98}}$ & 89.67$_{\text{1.91}}$ & 91.77$_{\text{1.62}}$ & 88.93 
& 74.13$_{\text{3.46}}$ & 80.58$_{\text{1.05}}$ & 80.28$_{\text{4.50}}$ & 65.06$_{\text{5.08}}$ & 60.32$_{\text{5.28}}$ & 72.07 \\
& F1  
& 81.84$_{\text{8.89}}$ & 90.79$_{\text{5.57}}$ & 89.89$_{\text{4.14}}$ & 89.63$_{\text{1.97}}$ & 91.76$_{\text{1.63}}$ & 88.78 
& 73.25$_{\text{2.85}}$ & 80.17$_{\text{1.26}}$ & 79.53$_{\text{5.06}}$ & 60.84$_{\text{7.63}}$ & 52.96$_{\text{9.04}}$ & 69.35 \\
\midrule
\multirow{2}{*}{PeCoLA$^\dag$} 
& Acc 
& 88.67$_{\text{1.74}}$ & 96.04$_{\text{0.41}}$ & 91.67$_{\text{0.97}}$ & 91.81$_{\text{0.67}}$ & 92.21$_{\text{0.85}}$ & 92.08 
& 88.83$_{\text{4.18}}$ & 88.32$_{\text{3.06}}$ & 87.87$_{\text{3.61}}$ & 67.73$_{\text{1.06}}$ & 61.51$_{\text{1.22}}$ & 78.85 \\
& F1  
& 88.57$_{\text{1.81}}$ & 96.02$_{\text{0.45}}$ & 91.65$_{\text{0.98}}$ & 91.80$_{\text{0.67}}$ & 92.48$_{\text{0.95}}$ & 92.10
& 88.68$_{\text{4.36}}$ & 88.21$_{\text{3.21}}$ & 87.69$_{\text{3.80}}$ & 65.28$_{\text{2.26}}$ & 55.30$_{\text{1.96}}$ & 77.03 \\
\midrule
\multirow{2}{*}{EAGLE$^\dag$} 
& Acc 
& 88.08$_{\text{3.22}}$ & 95.80$_{\text{0.62}}$ & 92.08$_{\text{1.15}}$ & 91.36$_{\text{1.40}}$ & 90.47$_{\text{4.10}}$ & 91.56 
& 78.54$_{\text{1.78}}$ & 82.31$_{\text{4.58}}$ & 82.84$_{\text{4.43}}$ & 65.85$_{\text{3.10}}$ & 57.36$_{\text{3.01}}$ &  73.38\\
& F1  
& 89.25$_{\text{2.61}}$ & 95.69$_{\text{0.66}}$  & 92.39$_{\text{1.03}}$ & 91.64$_{\text{1.12}}$ & 91.17$_{\text{3.27}}$  &  92.03
& 82.13$_{\text{1.24}}$  & 84.70$_{\text{3.24}}$  & 85.30$_{\text{3.26}}$  & 74.13$_{\text{1.74}}$ & 69.97$_{\text{1.41}}$  & 79.25 \\
\midrule
\multirow{2}{*}{ImBD} 
& Acc 
& 88.77$_{\text{0.95}}$ & 50.0$_{\text{0.20}}$ & 74.87$_{\text{1.72}}$ & 74.85$_{\text{1.46}}$ & 89.71$_{\text{0.47}}$ & 75.64 
& 88.65$_{\text{1.54}}$ & 77.35$_{\text{1.52}}$ & 84.37$_{\text{3.49}}$ & 54.66$_{\text{0.83}}$ & 50.02$_{\text{0.20}}$ & 71.01 \\
& F1  
& 88.85$_{\text{1.22}}$ & 66.67$_{\text{1.56}}$ & 79.65$_{\text{1.06}}$ & 79.82$_{\text{0.88}}$ & 89.06$_{\text{0.59}}$ & 80.81 
& 89.78$_{\text{1.25}}$ & 81.49$_{\text{1.00}}$ & 86.52$_{\text{2.66}}$ & 68.70$_{\text{0.37}}$ & 66.66$_{\text{0.32}}$ & 78.63 \\
\midrule
\multirow{2}{*}{MoSEs} 
& Acc 
& 86.39$_{\text{0.63}}$ & 87.87$_{\text{0.54}}$ & 86.30$_{\text{0.39}}$ & 82.09$_{\text{1.58}}$ & 91.17$_{\text{2.23}}$ & 86.76 
& 86.90$_{\text{1.78}}$ & 79.99$_{\text{2.90}}$ & 83.06$_{\text{2.17}}$ & 71.63$_{\text{1.86}}$ & 61.55$_{\text{2.24}}$ & 76.63 \\
& F1  
& 86.14$_{\text{0.64}}$ & 88.60$_{\text{0.58}}$  & 86.16$_{\text{0.47}}$ & 82.96$_{\text{1.89}}$ & 91.22$_{\text{2.64}}$  & 87.02 
& 86.61$_{\text{1.64}}$  & 78.98$_{\text{2.64}}$  & 82.14$_{\text{2.02}}$  & 72.12$_{\text{1.61}}$ & 48.35$_{\text{2.44}}$  & 73.64 \\
\midrule
\rowcolor{rowblue}
\multicolumn{14}{c}{\textbf{\textit{Domain generalization methods}}} \\
\midrule
\multirow{2}{*}{MSCL$^\dag$} 
& Acc 
& 87.50$_{\text{1.91}}$ & 95.87$_{\text{0.99}}$ & 91.60$_{\text{1.85}}$ & 90.33$_{\text{0.95}}$ & 92.71$_{\text{1.00}}$ & 91.60 
& 81.56$_{\text{5.8}}$ & 83.51$_{\text{3.08}}$ & 82.46$_{\text{4.21}}$ & 69.13$_{\text{3.72}}$ & 61.53$_{\text{2.39}}$ & 75.64 \\
& F1  
& 88.12$_{\text{1.78}}$ & 95.76$_{\text{1.04}}$ & 91.86$_{\text{1.59}}$ & 90.39$_{\text{0.72}}$ & 92.74$_{\text{0.99}}$ & 91.77
& 84.30$_{\text{4.29}}$ & 85.35$_{\text{2.17}}$ & 84.58$_{\text{3.09}}$ & 75.74$_{\text{2.17}}$ & 71.94$_{\text{1.16}}$ & 80.38 \\
\midrule
\multirow{2}{*}{TACIT} 
& Acc 
& 85.23$_{\text{0.78}}$ & 94.13$_{\text{0.77}}$ & 86.40$_{\text{0.87}}$ & 87.89$_{\text{0.67}}$ & 88.36$_{\text{0.96}}$ & 88.40 
& 84.88$_{\text{2.52}}$ & 81.09$_{\text{1.55}}$ & 87.22$_{\text{1.78}}$ & 64.73$_{\text{1.63}}$ & 57.99$_{\text{2.60}}$ & 75.18 \\
& F1  
& 85.76$_{\text{0.57}}$ & 94.18$_{\text{0.72}}$ & 87.09$_{\text{0.59}}$ & 88.34$_{\text{0.50}}$ & 88.79$_{\text{0.78}}$ & 88.83 
& 86.62$_{\text{1.97}}$ & 83.41$_{\text{1.05}}$ & 88.22$_{\text{1.38}}$ & 73.08$_{\text{0.82}}$ & 70.04$_{\text{1.23}}$ & 80.27 \\
\midrule
\multirow{2}{*}{MGT-Prism$^\dag$} 
& Acc 
& 88.76$_{\text{2.63}}$ & 96.28$_{\text{1.13}}$ & 91.81$_{\text{1.90}}$ & 91.14$_{\text{1.34}}$ & 91.93$_{\text{2.06}}$ & 91.98
& 89.08$_{\text{1.94}}$ & 87.34$_{\text{3.64}}$ & 88.58$_{\text{2.95}}$ & 70.53$_{\text{1.32}}$ & 65.47$_{\text{2.22}}$ &  80.20 \\
& F1  
& 89.02$_{\text{2.50}}$ & 96.23$_{\text{1.11}}$ & 92.57$_{\text{0.81}}$ & 91.16$_{\text{1.27}}$ & 92.06$_{\text{1.86}}$ & 92.21 
& 88.14$_{\text{3.59}}$ & 88.39$_{\text{2.85}}$ & 89.48$_{\text{2.34}}$ & 74.34$_{\text{0.95}}$ & \textbf{73.56$_{\text{1.20}}$} & 82.78 \\
\midrule
\multirow{2}{*}{\modelname{}$^\dag$} 
& Acc 
& \textbf{91.70$_{\text{0.74}}$} & \textbf{96.30$_{\text{0.28}}$} & \textbf{93.15$_{\text{1.12}}$} & \textbf{93.09$_{\text{0.52}}$} & \textbf{92.90$_{\text{0.67}}$} & \textbf{93.43} 
& \textbf{92.13$_{\text{2.28}}$} & \textbf{91.79$_{\text{0.95}}$} & \textbf{93.35$_{\text{1.42}}$} & \textbf{72.63$_{\text{1.06}}$} & 62.43$_{\text{0.98}}$ & \textbf{82.46} \\
& F1  
& \textbf{91.67$_{\text{0.93}}$} & \textbf{96.35$_{\text{0.54}}$} & \textbf{93.20$_{\text{0.95}}$} & \textbf{92.94$_{\text{0.65}}$} & \textbf{93.28$_{\text{0.53}}$} & \textbf{93.49} 
& \textbf{92.63$_{\text{1.88}}$} & \textbf{92.02$_{\text{0.79}}$} & \textbf{94.39$_{\text{1.67}}$} & \textbf{76.89$_{\text{1.06}}$} & 72.59$_{\text{0.10}}$ & \textbf{85.70} \\
\bottomrule
\end{tabular}
}
\caption{Comparison of \modelname{} and baseline methods in MGT detection under 10 test domains, using Accuracy and F1-score (\%) as evaluation metrics. 
Each result represents the average over 5 runs with different random seeds, and the subscript denotes the standard deviation. 
Results of metric-based methods are deterministic, and thus we omit standard deviations for them. 
$\dag$ indicates that the method uses RoBERTa-base as the backbone encoder.
The best scores are highlighted in \textbf{bold}.}
\label{main_exp}
  \vspace{-10pt}
\end{table*}


\fakeparagraph{Test-Time Expert Aggregation}
During inference, unseen-domain inputs often exhibit mixed stylistic or generation patterns~\cite{zhou2024cycle, dann2026principled}, rendering reliance on a single source-conditioned pathway suboptimal for robust detection. 
To address this, we leverage the trained policy to dynamically coordinate multiple expert pathways for each instance. 
Given an input $x$ encoded as $\mathbf{s}$, the policy network $\pi_{\theta_2}(\cdot \mid \mathbf{s})$ outputs a utility distribution over all candidate pathways. 
The top-$m$ pathways with the highest policy probabilities are selected for collaborative prediction, and their raw classification logits are aggregated via a confidence-weighted summation:
\begin{equation}
\small
P(\hat{y} \mid x) = \mathrm{softmax}\left( \sum_{j=1}^{m} \pi_{\theta_2}(a_j \mid s)\cdot z(\hat{y}\mid a_j, x) \right), \label{eq:final}
\end{equation}

where $\mathbf{z}(\hat{y} \mid a_j, x)$ denotes the pre-softmax logits produced by the $j$-th selected pathway.
By fusing complementary discriminative signals from multiple high-utility experts, \modelname{} achieves instance-adaptive expert orchestration, ensuring robust generalization to previously unseen domains.

\section{Experiment}


\subsection{Experiment Settings}
To demonstrate the effectiveness of \modelname{}, we use \textit{XSum}, \textit{HellaSwag}, \textit{SQuAD}, \textit{Yelp}, and \textit{Sci} datasets from the MAGE benchmark \cite{li-etal-2024-mage} as source domains for training the detector, and evaluate it on five unseen target domains (\textit{CMV}, \textit{ELI5}, \textit{WP}, \textit{TLDR}, \textit{ROCT}) to validate its performance in the \taskname setting.
Moreover, we assess the model under an IND setting where the training and test data come from the same domains. 
Comprehensive details of the datasets and key hyperparameters are provided in \appref{A.1} and \appref{app:hyper}, while the implementation details of our method are presented in \appref{A.2}.

\subsection{Comparison Models}

We compare \modelname{} with eleven baselines, including eight methods tailored for MGT detection and three DG approaches adapted to the detection task. 
\textbf{Metric-based methods}\footnote{For threshold-based methods, we follow the likelihood computation protocol of \citet{mireshghallah2023smaller}, using GPT-2 Small (124M) due to its stronger generalization in zero-shot detection.} include Fast-DetectGPT \cite{bao2024fastdetectgpt}, and Binoculars \cite{hans2024spotting}. \textbf{Model-based methods} include RoBERTa-base \cite{liu2019roberta}, Ghostbusters \cite{verma2024ghostbuster}, PeCoLA \cite{liu2024does}, EAGLE \cite{bhattacharjee2024eagle}, ImBD \cite{chen2025imitate}, and MoSEs \cite{wu2025moses}. \textbf{Domain generalization methods} include MSCL \cite{tan2022domain}, TACIT \cite{song2024tacit}, and  MGT-Prism \cite{liu2025mgt}.
\appref{B} describes all methods details.

\subsection{Performance Comparison}

We compare \modelname{} with existing MGT detection methods under both IND-MGT and \taskname settings, with results reported in \tabref{main_exp}. \modelname{} achieves the best average performance across all settings, outperforming the strongest baseline by \textbf{1.28\%} in F1-score and \textbf{1.35\%} in Accuracy on IND-MGT, and by \textbf{2.92\%} in F1-score and \textbf{2.26\%} in Accuracy on \taskname. These consistent improvements further support our pilot observations that effective generalization requires avoiding the dilution of domain-local cues during training and aligning source-knowledge selection with the detection objective at inference.
Compared with methods following the DIG paradigm, \ie MGT-Prism, \modelname{} improves the average F1-score by \textbf{1.28\%} on IND-MGT and \textbf{2.92\%} on \taskname. This shows that DMoE helps reduce knowledge dilution in DIG by using domain-specific experts to keep useful domain-local cues and shared experts to learn common machinization patterns. Compared with methods following the DMG paradigm, \ie MoSEs, \modelname{} further improves the average F1-score by \textbf{6.47\%} on IND-MGT and \textbf{12.06\%} on \taskname. This result shows that RL-based routing helps reduce routing misalignment in DMG by selecting and combining expert paths based on instance-level detection rewards instead of relevance matching.
Moreover, on challenging short-text domains, \ie \textit{TLDR} and \textit{ROCT}, where all methods struggle with short inputs, \modelname{} still improves the average F1-score over the strongest baseline by \textbf{0.79\%}, showing stable short-text detection performance.

\subsection{Ablation Study}
We conduct ablation studies on two key components of \modelname{}: the disentangled MoE architecture and the RL-based routing strategy. The effects of input length and inference latency are further discussed in \appref{app:length} and \ref{app:latency}.

\fakeparagraph{Expert Type Ablation}
We evaluate the effectiveness of the disentangled expert design by removing either the domain-specific or shared experts. Results are reported in \tabref{tab_ablation_combined}. 
\textbf{\textit{(i) base}} removes all experts and uses only RoBERTa-base \cite{liu2019roberta} with a task-specific classifier. 
\textbf{\textit{(ii) w/o. domain-specific expert}} keeps only shared experts, reducing the model to a conventional MMoE-style architecture \cite{ma2018modeling}. 
\textbf{\textit{(iii) w/o. shared expert}} retains only domain-specific experts. 
All variants are trained on the same data and evaluated on five unseen target domains.
Removing either expert type consistently degrades performance, showing that domain-specific experts capture domain-sensitive features while shared experts learn transferable representations. Their combination improves robustness under domain shift.

\fakeparagraph{Routing Strategy Ablation}
We further compare different routing strategies during inference (\tabref{tab_ablation_combined}): 
\textbf{\textit{(i) DMoE + oracle label}} uses oracle domain labels and uniformly aggregates all shared experts for OOD inputs; 
\textbf{\textit{(ii) DMoE + random label}} randomly assigns domain labels at test time; 
\textbf{\textit{(iii) DMoE + classifier}} uses a separately trained RoBERTa-base classifier for domain prediction.

Results show that these alternatives either rely on unrealistic supervision, unstable routing, or inaccurate domain prediction under distribution shift. In contrast, our policy model learns instance-adaptive routing from task rewards, leading to more effective expert selection and better generalization.

\begin{table}[t]
\centering
\resizebox{\linewidth}{!}{%
\small
\setlength{\tabcolsep}{3pt}
\begin{tabular}{lccccc}
\toprule
\textbf{Method} & \textbf{CMV} & \textbf{ELI5} & \textbf{WP} & \textbf{TLDR} & \textbf{ROCT} \\
\midrule
\multicolumn{6}{c}{\textbf{\textit{Expert Type Ablation}}} \\
\cmidrule(lr){1-6}
base (no experts)      & 86.03 & 85.45 & 83.68 & 74.46 & 69.74 \\
w/o domain-specific    & 89.24 & 90.14 & 91.46 & 75.64 & 70.86 \\
w/o domain-shared      & 89.62 & 90.02 & 91.52 & 76.40 & 71.37 \\
\modelname{}           & \textbf{92.63} & \textbf{92.02} & \textbf{94.39} & \textbf{76.89} & \textbf{72.59} \\
\midrule
\multicolumn{6}{c}{\textbf{\textit{Routing Strategy Ablation}}} \\
\cmidrule(lr){1-6}
DMoE + oracle label     & 92.27 & 91.86 & 93.97 & 76.10 & 71.32 \\
DMoE + random label     & 91.30 & 90.77 & 92.47 & 75.48 & 71.11 \\
DMoE + classifier     & 91.41 & 90.98 & 91.05 & 75.85 & 71.45 \\
\modelname{}           & \textbf{92.63} & \textbf{92.02} & \textbf{94.39} & \textbf{76.89} & \textbf{72.59} \\
\bottomrule
\end{tabular}
}
\caption{
\textbf{Ablation studies on expert types and routing strategies.} 
\textit{Top:} Comparison of expert configurations by removing domain-specific or domain-shared experts. 
\textit{Bottom:} Comparison of routing strategies, including oracle and random label variants.
}
\vspace{-0.3cm}
\label{tab_ablation_combined}
\end{table}

\subsection{Discussions and Analysis}

\begin{figure}[h]
  \centering
  \includegraphics[width=\linewidth]{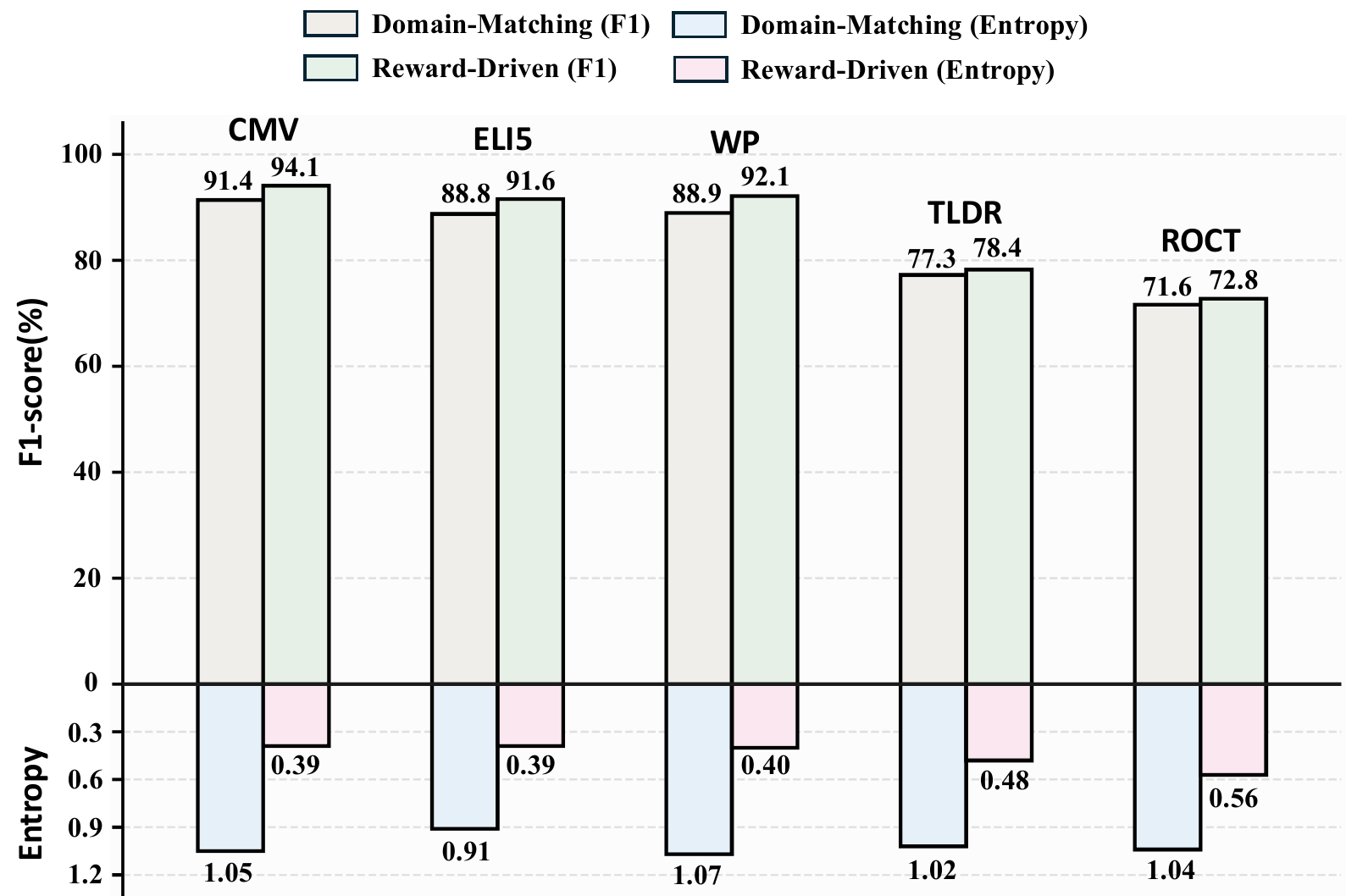}
    \caption{
    \textbf{Comparison of routing strategies on OOD datasets.}
    F1 scores and expert-pathway routing entropy are reported for domain-matching-based and reward-driven routing across five unseen-domain datasets.
    }
  \label{fig:router} 
\end{figure}

\fakeparagraph{Analysis on Routing Uncertainty} To verify the necessity of reward-driven routing, we compare RL-based routing with domain-matching routing on five unseen target domains. Domain-matching routing selects expert pathways according to domain association, which is not directly aligned with the final detection objective. Under OOD shifts, this mismatch leads to more uncertain routing and dispersed expert aggregation. As shown in Figure~\ref{fig:router}, RL-based routing improves the F1 score from 83.59 to 85.76 and reduces the routing entropy from 1.02 to 0.45, averaged over the five datasets. This confirms that optimizing routing with detection rewards enables \modelname{} to select more reliable expert pathways and achieve better generalization.

\fakeparagraph{Analysis on Robustness}
To further evaluate the robustness of our approach, we follow the experimental setting of \cite{chen2025imitate} and apply three paraphrasing attacks, namely Expand, Polish, and Rewrite, to the XSum dataset. As shown in \tabref{tab_paraphrasing_attack}, DEER consistently outperforms both model-based and metric-based baselines under all three attack settings. In particular, metric based methods experience a substantial drop in performance under this setup, indicating a strong reliance on shallow lexical cues that are easily disrupted by paraphrasing. Compared with the strongest baseline MGT-Prism, DEER achieves an average improvement of \textbf{1.07\%} in F1 score, demonstrating that our method maintains strong robustness while preserving good generalization ability. Additional robustness results are reported in \appref{app:robust}.

\begin{table}[h]
\centering
\small
\setlength{\tabcolsep}{5pt}
\begin{tabular}{lcccc}
\toprule
\textbf{Method} & \textbf{Expand} & \textbf{Polish} & \textbf{Rewrite} & \textbf{Avg.} \\
\midrule
Fast-DetectGPT & 67.55 & 55.07 & 38.06 & 53.56 \\
Binoculars     & 61.18 & 42.67 & 24.75  & 42.87 \\
MoSEs        & 65.75 & 65.28 & 60.59 & 63.87 \\
MGT-Prism       & 82.21 & 77.54 & 65.43 & 75.06 \\
\modelname{}    & \textbf{83.44} & \textbf{78.75} & \textbf{66.21} & \textbf{76.13} \\
\bottomrule
\end{tabular}
\caption{
\textbf{Robustness Evaluation Performance.}
}
\label{tab_paraphrasing_attack}
\end{table}

\fakeparagraph{Modular Expansion for Incremental Domain Adaptation}
To demonstrate the practicality of our disentangled architecture, we evaluate on an incremental deployment scenario where the model adapts to a new, unseen out-of-distribution (OOD) domain. We instantiate a dedicated domain-specific expert and fine-tune only this expert and the shared experts, keeping all existing experts frozen. 

\begin{figure}[ht]
\centering
\includegraphics[width=0.9\linewidth]{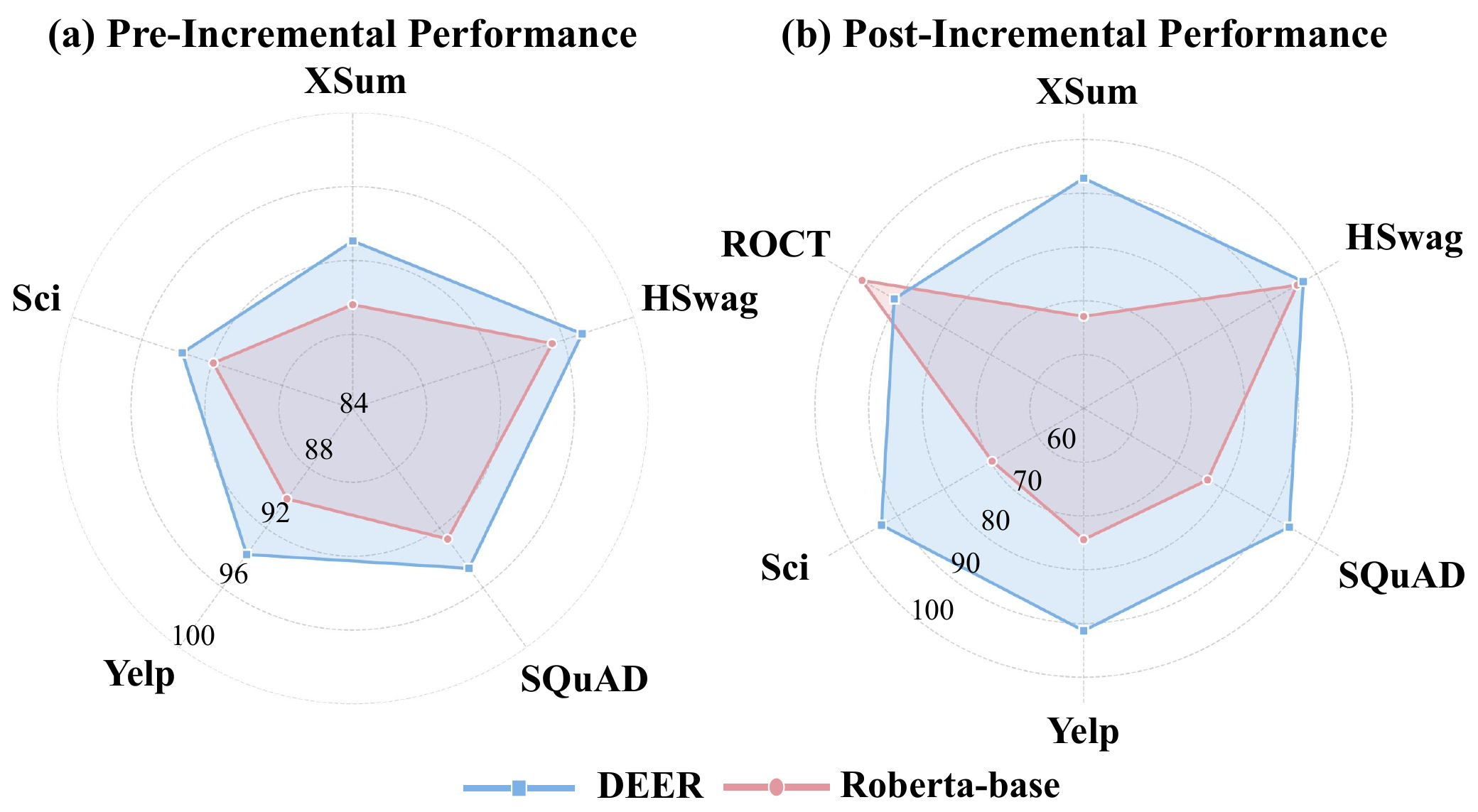} 
\caption{\textbf{Pre- and Post-incremental performance.} 
    (a) Result across source domains before incremental adaptation. 
    (b) Result after adapting to a new unseen domain, evaluated on both source and target domains.
    }
\label{fig:incre_performance}
\end{figure}

\begin{table}[ht]
\centering
\resizebox{\linewidth}{!}{%
\small
\setlength{\tabcolsep}{3pt}
\begin{tabular}{lccc}
\toprule
\textbf{Metric} & \textbf{RoBERTa-base} & \textbf{DEER} & \textbf{↓ Saving (\%)} \\
\midrule
Trainable Params (M) & 124.6 & 13.0 & \textbf{89.6\%} \\
GPU Memory (GB)      & 14.9  & 7.6  & \textbf{49.0\%} \\
Training Time (s)    & 320   & 189  & \textbf{40.9\%} \\
\bottomrule
\end{tabular}
}
\caption{\textbf{Efficiency comparison in the incremental OOD adaptation experiment.} }
\label{tab:efficiency}
  \vspace{-7pt}
\end{table}

As illustrated in \tabref{tab:efficiency} and \figref{fig:incre_performance}, our method achieves compelling performance comparable to fully fine-tuned models for new domain adaptation with negligible computational overhead. Compared with RoBERTa-base, DEER drastically reduces trainable parameters, GPU memory consumption and training cost, achieving up to \textbf{89.6\%} parameter reduction and \textbf{40.9\%} training time saving.
More importantly, vanilla RoBERTa-base obtains limited gains on new domains at the expense of degraded performance on existing source domains, while our method sustains stable performance across all source domains.  DEER supports efficient, stable and scalable adaptation to evolving data distributions.


\section{Conclusion}

In this work, we propose \modelname{}, a novel framework for robust machine-generated text detection under domain shift.
Our pilot studies reveal two key limitations of existing methods, \ie knowledge dilution caused by unified multi-domain training and routing misalignment introduced by similarity-based retrieval. 
To address these issues, \modelname{} introduces a Disentangled Mixture-of-Experts architecture with reinforcement learning-based routing, enabling adaptive expert selection while disentangling domain-specific and domain-invariant knowledge to preserve both discriminative features and transferable machinization patterns.
Experiments on ten diverse domains show that \modelname{} consistently outperforms sota detectors in both in-domain and out-of-distribution scenarios, while remaining robust to paraphrasing and style-based adversarial attacks. 
Overall, this work highlights the value of disentangled representations and adaptive expert routing for reliable open-world MGT detection.


\section*{Limitations}

Despite the promising results achieved by \modelname{} for generalizable MGT detection, there are three primary limitations to this study. \textbf{First}, the disentangled MoE training stage depends on domain labels for expert specialization, restricting its use when domain annotations are unavailable. \textbf{Second}, \modelname{} adopts a fixed expert capacity across all domains, and thus cannot automatically allocate more capacity to harder domains or reduce redundancy for simpler ones. \textbf{Third}, the RL-based routing mechanism operates over a predefined set of source domains, and its scalability and optimization stability under a large or evolving domain space remain underexplored. While our results are promising, extending \modelname{} to more realistic, dynamic domain scenarios is an important next step toward broader deployment.

\section*{Ethics Statement}

DEER aims to advance responsible MGT identification by improving robustness and generalization under realistic deployment settings where domain cues may be unavailable, noisy, or continuously evolving. We stress that detection scores should be interpreted as supportive evidence rather than sole grounds for consequential decisions, and we explicitly discourage any misuse of our findings to evade or undermine detection systems. All evaluations are conducted on publicly available datasets and do not involve collecting or releasing private or sensitive user data. More broadly, we hope this study motivates future research toward continual, incrementally updatable MGT detectors that can progressively adapt and strengthen in dynamic, non-stationary environments as new generators and domains emerge.

\bibliography{custom}

@misc{openai2025gpt5,
    author = {OpenAI},
    title = {GPT-5 System Card},
    year = {2025},
    url = {https://openai.com/index/gpt-5-system-card/},
}

@article{guo2025deepseek,
  title={Deepseek-r1: Incentivizing reasoning capability in llms via reinforcement learning},
  author={Guo, Daya and Yang, Dejian and Zhang, Haowei and Song, Junxiao and Zhang, Ruoyu and Xu, Runxin and Zhu, Qihao and Ma, Shirong and Wang, Peiyi and Bi, Xiao and others},
  journal={arXiv preprint arXiv:2501.12948},
  year={2025}
}

@misc{Anthropic2026Claude,
  title={Claude4.7},
  author={Anthropic},
  howpublished = {Website},
  url = {https://claude.ai/},
  year={2026}
}

@article{liu2024does,
  title={Does detectgpt fully utilize perturbation? bridging selective perturbation to fine-tuned contrastive learning detector would be better},
  author={Liu, Shengchao and Liu, Xiaoming and Wang, Yichen and Cheng, Zehua and Li, Chengzhengxu and Zhang, Zhaohan and Lan, Yu and Shen, Chao},
  journal={arXiv preprint arXiv:2402.00263},
  year={2024}
}

@inproceedings{li-etal-2024-mage,
    title = "{MAGE}: Machine-generated Text Detection in the Wild",
    author = "Li, Yafu  and
      Li, Qintong  and
      Cui, Leyang  and
      Bi, Wei  and
      Wang, Zhilin  and
      Wang, Longyue  and
      Yang, Linyi  and
      Shi, Shuming  and
      Zhang, Yue",
    editor = "Ku, Lun-Wei  and
      Martins, Andre  and
      Srikumar, Vivek",
    booktitle = "Proceedings of the 62nd Annual Meeting of the Association for Computational Linguistics (Volume 1: Long Papers)",
    month = aug,
    year = "2024",
    address = "Bangkok, Thailand",
    publisher = "Association for Computational Linguistics",
    url = "https://aclanthology.org/2024.acl-long.3/",
    doi = "10.18653/v1/2024.acl-long.3",
    pages = "36--53"
}

@inproceedings{chen2025imitate,
  title={Imitate Before Detect: Aligning Machine Stylistic Preference for Machine-Revised Text Detection},
  author={Chen, Jiaqi and Zhu, Xiaoye and Liu, Tianyang and Chen, Ying and Xinhui, Chen and Yuan, Yiwen and Leong, Chak Tou and Li, Zuchao and Tang, Long and Zhang, Lei and others},
  booktitle={Proceedings of the AAAI Conference on Artificial Intelligence},
  volume={39},
  number={22},
  pages={23559--23567},
  year={2025}
}

@misc{li2025ironsharpensirondefending,
      title={Iron Sharpens Iron: Defending Against Attacks in Machine-Generated Text Detection with Adversarial Training}, 
      author={Yuanfan Li and Zhaohan Zhang and Chengzhengxu Li and Chao Shen and Xiaoming Liu},
      year={2025},
      eprint={2502.12734},
      archivePrefix={arXiv},
      primaryClass={cs.CR},
      url={https://arxiv.org/abs/2502.12734}, 
}

@article{bhattacharjee2023conda,
  title={Conda: Contrastive domain adaptation for ai-generated text detection},
  author={Bhattacharjee, Amrita and Kumarage, Tharindu and Moraffah, Raha and Liu, Huan},
  journal={arXiv preprint arXiv:2309.03992},
  year={2023}
}

@article{hans2024spotting,
  title={Spotting llms with binoculars: Zero-shot detection of machine-generated text},
  author={Hans, Abhimanyu and Schwarzschild, Avi and Cherepanova, Valeriia and Kazemi, Hamid and Saha, Aniruddha and Goldblum, Micah and Geiping, Jonas and Goldstein, Tom},
  journal={arXiv preprint arXiv:2401.12070},
  year={2024}
}

@inproceedings{
wang2023seqxgpt,
title={Seq{XGPT}: Sentence-Level {AI}-Generated Text Detection},
author={Pengyu Wang and Linyang Li and Ke Ren and Botian Jiang and Dong Zhang and Xipeng Qiu},
booktitle={The 2023 Conference on Empirical Methods in Natural Language Processing},
year={2023},
url={https://openreview.net/forum?id=uemYdRTVvP}
}

@inproceedings{liu-etal-2023-coco,
  title={CoCo: Coherence-Enhanced Machine-Generated Text Detection Under Low Resource With Contrastive Learning},
  author={Liu, Xiaoming and Zhang, Zhaohan and Wang, Yichen and Pu, Hang and Lan, Yu and Shen, Chao},
  booktitle={Proceedings of the 2023 Conference on Empirical Methods in Natural Language Processing},
  pages={16167--16188},
  year={2023}
}

@article{shum2023automatic,
  title={Automatic Prompt Augmentation and Selection with Chain-of-Thought from Labeled Data},
  author={Shum, KaShun and Diao, Shizhe and Zhang, Tong},
  journal={arXiv preprint arXiv:2302.12822},
  year={2023}
}

@article{liu2019roberta,
  title={Roberta: A robustly optimized bert pretraining approach},
  author={Liu, Yinhan and Ott, Myle and Goyal, Naman and Du, Jingfei and Joshi, Mandar and Chen, Danqi and Levy, Omer and Lewis, Mike and Zettlemoyer, Luke and Stoyanov, Veselin},
  journal={arXiv preprint arXiv:1907.11692},
  year={2019}
}

@article{zellers2019defending,
  title={Defending against neural fake news},
  author={Zellers, Rowan and Holtzman, Ari and Rashkin, Hannah and Bisk, Yonatan and Farhadi, Ali and Roesner, Franziska and Choi, Yejin},
  journal={Advances in neural information processing systems},
  volume={32},
  year={2019}
}

@inproceedings{gehrmann2019gltr,
  title={GLTR: Statistical Detection and Visualization of Generated Text},
  author={Gehrmann, Sebastian and Strobelt, Hendrik and Rush, Alexander M},
  booktitle={Proceedings of the 57th Annual Meeting of the Association for Computational Linguistics: System Demonstrations},
  pages={111--116},
  year={2019}
}

@article{mireshghallah2023smaller,
  title={Smaller Language Models are Better Black-box Machine-Generated Text Detectors},
  author={Mireshghallah, Fatemehsadat and Mattern, Justus and Gao, Sicun and Shokri, Reza and Berg-Kirkpatrick, Taylor},
  journal={arXiv preprint arXiv:2305.09859},
  year={2023}
}

@inproceedings{
bao2024fastdetectgpt,
title={Fast-Detect{GPT}: Efficient Zero-Shot Detection of Machine-Generated Text via Conditional Probability Curvature},
author={Guangsheng Bao and Yanbin Zhao and Zhiyang Teng and Linyi Yang and Yue Zhang},
booktitle={The Twelfth International Conference on Learning Representations},
year={2024},
url={https://openreview.net/forum?id=Bpcgcr8E8Z}
}

@inproceedings{verma2024ghostbuster,
  title={Ghostbuster: Detecting Text Ghostwritten by Large Language Models},
  author={Verma, Vivek and Fleisig, Eve and Tomlin, Nicholas and Klein, Dan},
  booktitle={Proceedings of the 2024 Conference of the North American Chapter of the Association for Computational Linguistics: Human Language Technologies (Volume 1: Long Papers)},
  pages={1702--1717},
  year={2024}
}

@article{bhattacharjee2024eagle,
  title={Eagle: A domain generalization framework for ai-generated text detection},
  author={Bhattacharjee, Amrita and Moraffah, Raha and Garland, Joshua and Liu, Huan},
  journal={arXiv preprint arXiv:2403.15690},
  year={2024}
}

@inproceedings{ma2018modeling,
  title={Modeling task relationships in multi-task learning with multi-gate mixture-of-experts},
  author={Ma, Jiaqi and Zhao, Zhe and Yi, Xinyang and Chen, Jilin and Hong, Lichan and Chi, Ed H},
  booktitle={Proceedings of the 24th ACM SIGKDD international conference on knowledge discovery \& data mining},
  pages={1930--1939},
  year={2018}
}

@inproceedings{dai2021generalizable,
  title={Generalizable person re-identification with relevance-aware mixture of experts},
  author={Dai, Yongxing and Li, Xiaotong and Liu, Jun and Tong, Zekun and Duan, Ling-Yu},
  booktitle={Proceedings of the IEEE/CVF conference on computer vision and pattern recognition},
  pages={16145--16154},
  year={2021}
}

@article{zhong2022meta,
  title={Meta-dmoe: Adapting to domain shift by meta-distillation from mixture-of-experts},
  author={Zhong, Tao and Chi, Zhixiang and Gu, Li and Wang, Yang and Yu, Yuanhao and Tang, Jin},
  journal={Advances in Neural Information Processing Systems},
  volume={35},
  pages={22243--22257},
  year={2022}
}

@article{sutton1988learning,
  title={Learning to predict by the methods of temporal differences},
  author={Sutton, Richard S},
  journal={Machine learning},
  volume={3},
  pages={9--44},
  year={1988},
  publisher={Springer}
}

@article{qu2022hmoe,
  title={Hmoe: Hypernetwork-based mixture of experts for domain generalization},
  author={Qu, Jingang and Faney, Thibault and Wang, Ze and Gallinari, Patrick and Yousef, Soleiman and de Hemptinne, Jean-Charles},
  journal={arXiv preprint arXiv:2211.08253},
  year={2022}
}

@inproceedings{tan2022domain,
  title={Domain generalization for text classification with memory-based supervised contrastive learning},
  author={Tan, Qingyu and He, Ruidan and Bing, Lidong and Ng, Hwee Tou},
  booktitle={Proceedings of the 29th International Conference on Computational Linguistics},
  pages={6916--6926},
  year={2022}
}

@inproceedings{song2024tacit,
  title={Tacit: A target-agnostic feature disentanglement framework for cross-domain text classification},
  author={Song, Rui and Giunchiglia, Fausto and Li, Yingji and Tian, Mingjie and Xu, Hao},
  booktitle={Proceedings of the AAAI Conference on Artificial Intelligence},
  volume={38},
  number={17},
  pages={18999--19007},
  year={2024}
}

@article{jacobs1991adaptive,
  title={Adaptive mixtures of local experts},
  author={Jacobs, Robert A and Jordan, Michael I and Nowlan, Steven J and Hinton, Geoffrey E},
  journal={Neural computation},
  volume={3},
  number={1},
  pages={79--87},
  year={1991},
  publisher={MIT Press}
}

@article{ren2023pangu,
  title={Pangu-$\{$$\backslash$Sigma$\}$: Towards trillion parameter language model with sparse heterogeneous computing},
  author={Ren, Xiaozhe and Zhou, Pingyi and Meng, Xinfan and Huang, Xinjing and Wang, Yadao and Wang, Weichao and Li, Pengfei and Zhang, Xiaoda and Podolskiy, Alexander and Arshinov, Grigory and others},
  journal={arXiv preprint arXiv:2303.10845},
  year={2023}
}

@inproceedings{radwan2025feddg,
  title={FedDG-MoE: Test-Time Mixture-of-Experts Fusion for Federated Domain Generalization},
  author={Radwan, Ahmed and Soliman, Mahmoud and Abdelaziz, Omar and Shehata, Mohamed},
  booktitle={Proceedings of the Computer Vision and Pattern Recognition Conference},
  pages={1811--1820},
  year={2025}
}

@article{xu2024cbdmoe,
  title={CBDMoE: Consistent-but-Diverse Mixture of Experts for Domain Generalization},
  author={Xu, Fangbin and Chen, Dongyue and Jia, Tong and Deng, Shizhuo and Wang, Hao},
  journal={IEEE Transactions on Multimedia},
  volume={26},
  pages={9814--9824},
  year={2024},
  publisher={IEEE}
}

@article{chen2024lfme,
  title={Lfme: A simple framework for learning from multiple experts in domain generalization},
  author={Chen, Liang and Zhang, Yong and Song, Yibing and Shen, Zhiqiang and Liu, Lingqiao},
  journal={Advances in Neural Information Processing Systems},
  volume={37},
  pages={102919--102947},
  year={2024}
}

@article{liu2025mgt,
  title={MGT-Prism: Enhancing Domain Generalization for Machine-Generated Text Detection via Spectral Alignment},
  author={Liu, Shengchao and Liu, Xiaoming and Li, Chengzhengxu and Zhang, Zhaohan and Ma, Guoxin and Lan, Yu and Xiao, Shuai},
  journal={arXiv preprint arXiv:2508.13768},
  year={2025}
}

@inproceedings{wu2025moses,
  title={Moses: Uncertainty-aware ai-generated text detection via mixture of stylistics experts with conditional thresholds},
  author={Wu, Junxi and Wang, Jinpeng and Liu, Zheng and Chen, Bin and Hu, Dongjian and Wu, Hao and Xia, Shu-Tao},
  booktitle={Proceedings of the 2025 Conference on Empirical Methods in Natural Language Processing},
  pages={5797--5816},
  year={2025}
}

@article{guo2024biscope,
  title={BiScope: AI-generated Text Detection by Checking Memorization of Preceding Tokens},
  author={Guo, Hanxi and Cheng, Siyuan and Jin, Xiaolong and Zhang, Zhuo and Zhang, Kaiyuan and Tao, Guanhong and Shen, Guangyu and Zhang, Xiangyu},
  journal={Advances in Neural Information Processing Systems (NeurIPS)},
  volume={37},
  pages={104065--104090},
  year={2024}
}

@inproceedings{wen2025measure,
  title={Measure Domain's Gap: A Similar Domain Selection Principle for Multi-Domain Recommendation},
  author={Wen, Yi and Liu, Yue and Xu, Derong and Luo, Huishi and Jia, Pengyue and Wu, Yiqing and Wang, Siwei and Liang, Ke and Wang, Maolin and Wang, Yiqi and others},
  booktitle={Proceedings of the 31st ACM SIGKDD Conference on Knowledge Discovery and Data Mining V. 2},
  pages={3156--3167},
  year={2025}
}

@inproceedings{nicks2024language,
  title={Language model detectors are easily optimized against},
  author={Nicks, Charlotte and Mitchell, Eric and Rafailov, Rafael and Sharma, Archit and Manning, Christopher and Finn, Chelsea and Ermon, Stefano},
  booktitle={International Conference on Learning Representations},
  volume={2024},
  pages={7807--7826},
  year={2024}
}

@inproceedings{pedrotti2025stress,
  title={Stress-testing machine generated text detection: Shifting language models writing style to fool detectors},
  author={Pedrotti, Andrea and Papucci, Michele and Ciaccio, Cristiano and Miaschi, Alessio and Puccetti, Giovanni and Dell’Orletta, Felice and Esuli, Andrea},
  booktitle={Findings of the Association for Computational Linguistics: ACL 2025},
  pages={3010--3031},
  year={2025}
}

@article{dann2026principled,
  title={Principled model routing for unknown mixtures of source domains},
  author={Dann, Christoph and Mansour, Yishay and Marinov, Teodor Vanislavov and Mohri, Mehryar},
  journal={Advances in Neural Information Processing Systems},
  volume={38},
  pages={40626--40657},
  year={2026}
}

@inproceedings{zhou2024cycle,
  title={Cycle self-refinement for multi-source domain adaptation},
  author={Zhou, Chaoyang and Wang, Zengmao and Du, Bo and Luo, Yong},
  booktitle={Proceedings of the AAAI Conference on Artificial Intelligence},
  volume={38},
  number={15},
  pages={17096--17104},
  year={2024}
}

@article{zhang2021quantifying,
  title={Quantifying and improving transferability in domain generalization},
  author={Zhang, Guojun and Zhao, Han and Yu, Yaoliang and Poupart, Pascal},
  journal={Advances in Neural Information Processing Systems},
  volume={34},
  pages={10957--10970},
  year={2021}
}

@article{wong2026k,
  title={$ k $ NNProxy: Efficient Training-Free Proxy Alignment for Black-Box Zero-Shot LLM-Generated Text Detection},
  author={Wong, Kahim and Li, Kemou and Wu, Haiwei and Zhou, Jiantao},
  journal={arXiv preprint arXiv:2604.02008},
  year={2026}
}

\clearpage
\newpage
\appendix

\section*{Appendix}

\section{Experiment Setting Details}\label{A}



\subsection{Dataset Details}\label{A.1}

We evaluate our method on 10 domain-specific datasets curated from the MAGE benchmark. Specifically, \textit{Xsum}, \textit{HellaSwag}, \textit{SQuAD}, \textit{Yelp}, and \textit{Sci} are used as source domains for training and validation. All baselines and our model are trained on the same source training set and validated on the same held-out set to ensure a fair comparison. For testing, we construct 10 domain-specific test sets, each containing examples from a single domain. Based on these test sets, we define two evaluation settings: IND-MGT evaluates performance on the five seen source domains, while DG-MGT measures zero-shot generalization to five unseen target domains (\textit{CMV}, \textit{ELI5}, \textit{WP}, \textit{TLDR}, and \textit{ROCT}). This setup enables a comprehensive assessment of both cross-domain generalization and potential degradation in source-domain performance resulting from multi-domain training.
The detailed statistics of each dataset are summarized in Table~\ref{tab:dataset-stats}.

\begin{table}[h]
\centering
\small
\setlength{\tabcolsep}{3pt}
\begin{tabular*}{\linewidth}{@{\extracolsep{\fill}} c c c c c}
\toprule
\textbf{Dataset} & \textbf{\# Instances} & \textbf{H. Tokens} & \textbf{M. Tokens} & \textbf{\# Classes} \\
\midrule
\multicolumn{5}{c}{\textit{Training Set from Source Domains}} \\
\midrule
Train       & 10,000 & 202 & 312 & 2 \\
Validation  & 3,737  & 209 & 309 & 2 \\
\midrule
\multicolumn{5}{c}{\textit{Test Sets for IND-MGT Evaluation}} \\
\midrule
Xsum        & 1,587  & 489 & 226 & 2 \\
HellaSwag   & 1,575  & 32 & 327 & 2 \\
SQuAD       & 1,548  & 149 & 383 & 2 \\
Yelp        & 1,558  & 156 & 296 & 2 \\
Sci         & 1,280  & 205 & 305 & 2 \\
\midrule
\multicolumn{5}{c}{\textit{Test Sets for DG-MGT Evaluation}} \\
\midrule
CMV         & 4,800  & 350 & 421 & 2 \\
ELI5        & 4,800  & 334 & 361 & 2 \\
WP          & 4,800  & 670 & 474 & 2 \\
TLDR        & 4,800  & 95 & 261 & 2 \\
ROCT        & 4,800  & 51 & 372 & 2 \\
\bottomrule
\end{tabular*}
\caption{\textbf{Statistics of datasets used in our experiments.} 
The training and validation sets are drawn from five source domains, and all ten domains are used for evaluation.
\textbf{H. Tokens} and \textbf{M. Tokens} indicate the average token length of human-written and machine-generated texts, respectively.}
\label{tab:dataset-stats}
\end{table}

\subsection{Implementation Details}\label{A.2}
We initialize our model using \texttt{RoBERTa-base} as the backbone encoder. In the first stage, the Disentangled Mixture-of-Experts (DMoE) framework is trained for 30 epochs with a batch size of 16, using the AdamW optimizer with a learning rate of 2e-5. For each domain, we construct $m_1 = 5$ domain-specific experts, and globally include $m_2 = 6$ domain-shared experts shared across all domains. Each expert, whether domain-specific or domain-shared, is implemented as an independent two-layer MLP with ReLU activations. In the second stage, we train a policy network to simulate expert selection over an action space of five domain indices (corresponding to the training domains). The policy network consists of two linear layers with dimensions $w_1 \in \mathbb{R}^{768 \times 512}$ and $w_2 \in \mathbb{R}^{512 \times 5}$, and is optimized using AdamW with a learning rate of 1e-3 and weight decay $\varepsilon = 1\text{e}{-5}$. The policy is trained for 100 epochs with a batch size of 16. All experiments are conducted on a single NVIDIA A100 80GB GPUs.


\subsection{RL Training Details}\label{A.3}

\fakeparagraph{Normalized State Representation}
To enhance the generalization ability of the policy network across diverse domains, we employ \textit{dynamic feature normalization}. Specifically, we maintain a running estimate of the mean and standard deviation of the state representations during training, and normalize each state $s$ accordingly. This technique improves numerical stability, mitigates domain-induced distribution shifts, and facilitates more robust policy learning.

\fakeparagraph{Stabilized Reward Scaling}
To address instability caused by large inter-sample reward variance, we adopt \textit{batch-level reward normalization}. Specifically, rewards within each training batch are standardized to have zero mean and unit variance. This normalization ensures a consistent reward scale, stabilizes gradient updates, and helps prevent policy collapse or overfitting to outlier samples.

\subsection{Overall Procedure}\label{A.4}

In this section, we present the overall procedure of \modelname{}, which follows a two-stage training framework with test-time expert routing. In the first stage, DMoE learns disentangled representations by preserving domain-local detection cues with specialized experts and capturing transferable machinization patterns with shared experts. In the second stage, a policy network is optimized to estimate the task utility of each expert pathway using detection-driven rewards. During inference, \modelname{} only relies on the input text: the policy network selects the top-$m$ high-utility expert pathways, whose outputs are aggregated in an instance-adaptive manner for final prediction.
The complete procedure is summarized in Algorithm~\ref{alg:1}.

\begin{algorithm}[ht]
\small
\captionsetup{font=small}
\caption{Two-Stage Training and Inference Pipeline}
\label{alg:1}
\begin{algorithmic}[1]

  \STATE \textbf{Input:} Training set $\mathcal{D}_{\text{train}}$, testing set $\mathcal{D}_{\text{test}}$,
  a pre-trained text encoder $\mathrm{Encoder}(\cdot)$, a lightweight classification head
  $\mathcal{M}_{\theta_{1}}(\cdot)$,
  a set of domain-specific experts
  $\{\mathcal{E}_{\text{ds}}^{1}, \mathcal{E}_{\text{ds}}^{2}, \dots, \mathcal{E}_{\text{ds}}^{n}\}$,
  a set of domain-specific gate functions $\{G_1(\cdot), G_2(\cdot), \dots, G_n(\cdot)\}$,
  a domain-shared expert $\mathcal{E}_{\text{dc}}$, and a policy network
  $\pi_{\theta_{2}}(\cdot \mid \cdot)$
  \STATE \textbf{Output:} Predictions for test inputs

  \Stage{**** The First Training Stage ****}
  \STATE Initialize all learnable parameters and $epoch \leftarrow 0$
  \WHILE{$epoch < epoch_{max}$}
    \FOR{each $(x_{k}, y_{k}, d_{k})$ in $\mathcal{D}_{\text{train}}$}
      \STATE Encode text $x_{k}$ by $\mathrm{Encoder}(\cdot)$ as Eq.~(1)
      \STATE Get the weight vector $w_{k}$ using $G_k(\cdot)$ according to $d_k$ as Eq.~(2)
      \STATE Get final representation $H_k$ by $\mathcal{E}_{\text{ds}}^{k}$ and $\mathcal{E}_{\text{dc}}$ as Eq.~(3)
      \STATE Calculate the predicted label $\hat{y}_k$ by $\mathcal{M}_{\theta_{1}}(\cdot)$ and the cross-entropy loss
    \ENDFOR
    \STATE Update all learnable parameters with the cross-entropy loss
    \STATE $epoch \leftarrow epoch + 1$
  \ENDWHILE

  \Stage{**** The Second Training Stage ****}
  \STATE Freeze all learnable parameters except $\pi_{\theta_{2}}(\cdot \mid \cdot)$ and set $epoch \leftarrow 0$
  \WHILE{$epoch < epoch_{max}$}
    \FOR{each $(x_{i}, y_{i})$ in $\mathcal{D}_{\text{train}}$}
      \STATE Encode $x_{i}$ to get state $s_{i}$
      \STATE Sample action $a_i$ by $\pi_{\theta_{2}}(a_{i}\mid s_{i})$ as Eq.~(4)
      \STATE Calculate advantage-based reward $r_{\text{final}}$ by Eq.~(7)
      \STATE Store $(s_{i}, a_{i}, r_{\text{final}})$
    \ENDFOR
    \STATE Update $\pi_{\theta_{2}}(\cdot \mid \cdot)$ using REINFORCE with stored trajectories
    \STATE $epoch \leftarrow epoch + 1$
  \ENDWHILE

  \Stage{**** The Inference Stage ****}
  \STATE Freeze all learnable parameters
  \FOR{each $x$ in $\mathcal{D}_{\text{test}}$}
    \STATE Encode $x$ to get state $s$
    \STATE Compute top-$m$ domain probabilities by $\pi_{\theta_{2}}(a\mid s)$
    \STATE Aggregate top-$m$ expert pathways to obtain the final prediction $\hat{y}$ by Eq.~(8)
  \ENDFOR

\end{algorithmic}
\end{algorithm}

\section{Baseline Details}\label{B}

We compare DEER with eleven baselines, including nine methods tailored for MGT detection and two domain generalization approaches adapted to the detection task.

\noindent \textbf{\textit{Metric-based detectors}} make predictions by computing the log-probability scores from a generative language model and applying a fixed decision threshold.


\noindent \textbf{Fast-DetectGPT}, an optimized version of DetectGPT, it offering significant speedup  while maintaining or even surpassing the detection accuracy of DetectGPT.

\noindent \textbf{Binoculars} contrasts perplexity and cross-perplexity between two language models to detect machine-generated text without any fine-tuning.

\noindent \textbf{\textit{Model-based detectors}} adapt pre-trained language models(PLM) to the detection task by training on labeled data with a classification objective.

\noindent \textbf{RoBERTa}, a transformer-based model fine-tuned for binary classification.



\noindent \textbf{Ghostbusters} extracts token-level statistical features from weak LMs and trains a linear classifier for black-box detection.

\noindent \textbf{PeCoLA} applies token-aware perturbations and contrastive learning to better distinguish machine-generated text while preserving key semantics.

\noindent \textbf{EAGLE} integrates domain-adversarial training and contrastive learning for domain-generalized AI-generated text detection.

\noindent \textbf{ImBD} aligns a scoring model with machine-style preferences and detects machine-generated text via style-conditional probability curvature.

\noindent \textbf{MoSEs} models profession-specific writing styles via a mixture of stylistic experts and adopts conditional thresholding for uncertainty-aware MGT detection.

\noindent \textbf{\textit{Domain generalization methods}} aim to improve robustness under domain shift by learning domain-invariant representations, without relying on target domain data during training.

\noindent \textbf{MSCL} combines supervised contrastive learning with memory-based feature alignment to learn domain-invariant representations.

\noindent \textbf{TACIT} disentangles robust and non-robust features through a two-stage feature alignment framework to improve domain generalization.

\noindent \textbf{MGT-Prism} exploits frequency-domain features with spectral filtering and alignment to improve machine-generated text detection under domain shift.

\section{Efficiency of DEER}\label{C}

\subsection{Effect of Text Length}\label{app:length}

To evaluate the ability of \modelname{} to detect MGTs of varying lengths, we divide the test samples into segments of 100, 200, 300, and 400 tokens under the WP dataset. As shown in \tabref{tab_vary_text_len}, \modelname{} consistently outperforms all SOTA methods across different input lengths. Furthermore, we observe a steady improvement in F1-score as the input length increases, indicating that existing models generally struggle to detect short machine-generated texts due to the lack of sufficient discriminative signals.

\begin{table}[t]
\centering
\small
\setlength{\tabcolsep}{5pt}
\begin{tabular}{lccccc}
\toprule
\textbf{Method} & \textbf{100} & \textbf{200} & \textbf{300} & \textbf{400} & \textbf{Avg.} \\
\midrule
Binoculars     & 73.59 & 82.68 & 84.1 & 84.23 & 81.15 \\
Roberta-base   & 73.53 & 83.87 & 88.45 & 89.92 & 83.94 \\
MoSEs  & 74.15 & 83.72 & 85.66 & 86.63 & 82.54 \\
MGT-Prism  & 85.71 & 93.60 & 95.23 & 95.54 & 92.52 \\
\modelname{}   & \textbf{86.41} & \textbf{94.59} & \textbf{96.13} & \textbf{96.56} & \textbf{93.42} \\
\bottomrule
\end{tabular}
\caption{
\textbf{Performance across Varying Text Lengths.}
}
\label{tab_vary_text_len}
\end{table}



\subsection{Effect of Inference Latency}\label{app:latency}



To further evaluate efficiency, we report per-sample inference latency and GPU memory usage in Table~\ref{tab_inference_efficiency}. Metric-based detectors rely on large generative models for token-level likelihood evaluation, incurring substantial latency and memory overhead, while MoSEs suffers from high inference cost due to instance-wise test-time retrieval and adaptation. In contrast, MGT-Prism and \modelname{} adopt RoBERTa-base as a lightweight backbone, achieving significantly lower latency and memory footprint. Notably, though \modelname{} aggregates multiple experts via repeated forward passes, it avoids per-token scoring and test-time optimization, maintaining efficiency on par with MGT-Prism while delivering superior detection performance.

\begin{table}[h]
\centering
\small
\setlength{\tabcolsep}{10pt}
\renewcommand{\arraystretch}{1.1}
\begin{tabular}{l c c}
\toprule
\textbf{Method} & \textbf{Latency (ms)} & \textbf{Mem. (GB)} \\
\midrule

\rowcolor{rowblue}
\multicolumn{3}{c}{\textit{Metric-based Methods}} \\
Fast-DetectGPT  & 75.6 & 23.16 \\
Binoculars      & 200  & 27.18 \\

\midrule
\rowcolor{rowblue}
\multicolumn{3}{c}{\textit{Model-based Methods}} \\
MoSEs           & 1175 & 24.76 \\
MGT-Prism       & \textbf{13.4} & \underline{5.52} \\

\midrule
\rowcolor{rowblue}
\multicolumn{3}{c}{\textit{Our Method}} \\
\modelname{}    & \underline{27.4} & \textbf{5.21} \\

\bottomrule
\end{tabular}
\caption{
\textbf{Inference Efficiency Comparison.} 
Latency denotes the average inference time per sample, and Mem. represents the peak GPU memory usage during inference. 
All methods are evaluated under the same experimental settings.
}
\label{tab_inference_efficiency}
\end{table}

\section{Additional Experiments Results}\label{D}



\subsection{Analysis on Robustness}\label{app:robust}

To comprehensively evaluate robustness, we test \modelname{} under two complementary attacks: token-level perturbation attacks and DPO-based style alignment attacks, targeting local input noise and human-like stylistic shifts, respectively.

\fakeparagraph{Token-level Perturbation}
We first assess robustness against token-level post-hoc perturbations by applying repetition, deletion, and replacement operations to tokens in the ROCT test set. As shown in \tabref{tab_perturbation}, \modelname{} consistently outperforms both model-based and metric-based baselines across all perturbation strategies. Compared with the best-performing baseline, \modelname{} achieves an average absolute improvement of \textbf{2.41\%} in F1-score, demonstrating strong resilience to local input noise in addition to its cross-domain generalization ability.

\begin{table}[h]
\centering
\small
\setlength{\tabcolsep}{6pt}
\begin{tabular}{lcccc}
\toprule
\textbf{Method} & \textbf{Repeat} & \textbf{Delete} & \textbf{Replace} & \textbf{Avg.} \\
\midrule
Fast-DetectGPT & 69.62 & 69.29 & 56.95 & 65.29 \\
Binoculars     & 70.57 & 70.29 & 58.61 & 66.49 \\
MGT-Prism         & 64.76 & 72.18 & 63.93 & 66.96 \\
\modelname{}   & \textbf{72.35} & \textbf{78.92} & \textbf{66.44} & \textbf{72.57} \\
\bottomrule
\end{tabular}
\caption{
\textbf{Robustness under Perturbation Attacks.}
Average F1-scores (\%) on the ROCT dataset under three types of input perturbations.
}
\label{tab_perturbation}
\end{table}

\fakeparagraph{DPO-based Style Alignment Attack}
We further evaluate \modelname{} against DPO-based style alignment attacks, where generators are optimized to produce human-like MGTs that evade detection. Following related adversarial benchmarks~\cite{pedrotti2025stress, nicks2024language}, detectors are trained on clean data and evaluated on DPO-optimized adversarial test data. As shown in \tabref{tab_dpo_attack}, \modelname{} achieves the best performance on both domains, outperforming the strongest baseline by \textbf{1.33\%} on XSum and \textbf{7.16\%} on arXiv. These results indicate that \modelname{} remains robust even when test samples are explicitly optimized to weaken detector-relevant signals.

\begin{table}[h]
\centering
\small
\setlength{\tabcolsep}{8pt}
\renewcommand{\arraystretch}{1.12}
\begin{tabular*}{0.92\linewidth}{@{\extracolsep{\fill}}lccc}
\toprule
\textbf{Method} & \textbf{XSum} & \textbf{arXiv} & \textbf{Avg.} \\
\midrule
Entropy        & 27.28 & 48.70 & 37.99 \\
Log-Likelihood & 35.80 & 50.44 & 43.12 \\
LRR            & 36.80 & 50.30 & 43.55 \\
Fast-DetectGPT & 39.64 & 48.50 & 44.07 \\
Binoculars     & 46.70 & 50.24 & 48.47 \\
MGT-Prism      & 50.23 & 83.52 & 66.88 \\
\modelname{}   & \textbf{51.56} & \textbf{90.68} & \textbf{71.12} \\
\bottomrule
\end{tabular*}
\caption{
\textbf{DPO-based Style Alignment Attacks.}
Accuracy (\%) across XSum and Arxiv domains.
}

\label{tab_dpo_attack}
\end{table}

\subsection{Analysis on Hyperparameter}\label{app:hyper}

We analyze the impact of two critical hyperparameters in our framework: \textbf{\textit{i}}) the number of top-$m$ expert groups selected by the RL agent, and \textbf{\textit{ii}}) the number of sub-networks for domain-specific ($m_1$) and domain-shared ($m_2$) experts.

\begin{figure}[h]
\centering
\includegraphics[width=1.0\linewidth]{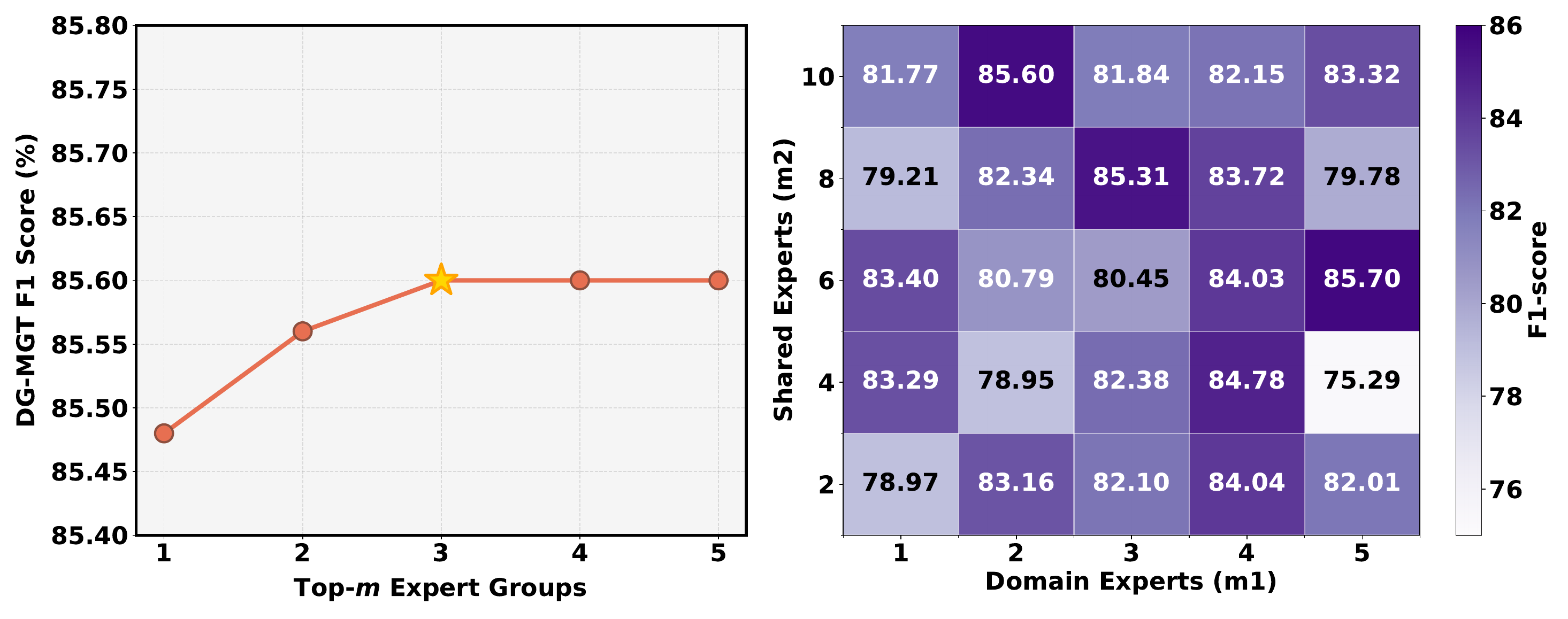} 
\caption{\textbf{Hyperparameter analysis on DG-MGT.} Left: performance variation with different values of top-$m$ expert group selection.
Right: F1-score heatmap under different configurations of shared and domain-specific experts. }
\label{fig:hyper-param}
\end{figure}

As shown in the left plot of Figure~\ref{fig:hyper-param}, increasing $m$ leads to a consistent improvement in DG-MGT F1 score, suggesting that aggregating expert groups from multiple domains facilitates cross-domain interaction and enhances generalization. However, performance gains plateau beyond $m{=}3$, indicating limited marginal benefits from additional experts. The RL agent addresses this by adaptively downweighting less informative groups, highlighting its ability to focus on useful domain knowledge.

In addition, we explore various configurations of $m_1$ and $m_2$ and adopt the best-performing setting as reported in the right heatmap of Figure~\ref{fig:hyper-param}.

\vspace{-0.2cm}

\subsection{Analysis on Scalability}\label{app:attack_scalability}
To examine the scalability of \modelname{} to emerging attacks, we study whether adversarial data can be incorporated as a new distribution through attack-specific modeling. 
We use DPO-optimized adversarial samples as the attack distribution and compare two training settings. 
In the \emph{Clean Trained} setting, detectors are trained only on the original clean data and evaluated directly on adversarial samples, measuring their baseline robustness to the attack. 
In the \emph{Merged Trained} setting, adversarial samples are added to the training set and detectors are retrained, measuring whether exposure to attack data improves robustness. 
For \modelname{}, the adversarial samples are treated as a distinct domain and assigned a dedicated attack-specific expert, which tests whether its modular design can model attack-specific patterns while avoiding entanglement with clean-domain representations.

\begin{table}[h]
\centering
\small
\setlength{\tabcolsep}{6pt}
\renewcommand{\arraystretch}{1.12}
\begin{tabular*}{0.95\linewidth}{@{\extracolsep{\fill}}lcc}
\toprule
\textbf{Method} & \textbf{Clean Trained} & \textbf{Merged Trained} \\
\midrule
Entropy        & 27.28 & 26.76 \\
Log-Likelihood & 35.80 & 37.40 \\
LRR            & 36.80 & 40.72 \\
Fast-DetectGPT & 39.64 & 50.00 \\
Binoculars     & 46.70 & 43.46 \\
MGT-Prism      & 50.23 & 82.79 \\
\modelname{}   & \textbf{51.56} & \textbf{87.84} \\
\bottomrule
\end{tabular*}
\caption{
\textbf{Scalability to Attack-Specific Experts.}
Accuracy (\%) on DPO-optimized adversarial test data with clean-only and merged training.
}
\label{tab_attack_expert}
\end{table}

As shown in \tabref{tab_attack_expert}, merging adversarial samples yields inconsistent gains for existing detectors, reflecting the distributional mismatch between clean and attack data. 
In contrast, \modelname{} benefits most from merged training, improving from 51.56\% to 87.84\% in accuracy and achieving the best overall performance, highlighting the advantage of dedicated attack-specific experts in adapting to new attack distributions while mitigating interference with clean-domain representations.






\begin{table*}[ht]
\small
\centering
\resizebox{\textwidth}{!}{
\setlength{\tabcolsep}{3pt}
\begin{tabular}{l c cccccc cccccc}
\toprule
\multirow{2}{*}{\textbf{Methods}} 
& \multirow{2}{*}{\textbf{Metric}} 
& \multicolumn{6}{c}{\textbf{IND-MGT}} 
& \multicolumn{6}{c}{\textbf{DG-MGT}} \\
\cmidrule(lr){3-8} \cmidrule(lr){9-14}
& 
& \textit{\textbf{XSum}} & \textit{\textbf{HSwag.}} & \textit{\textbf{SQuAD}} & \textit{\textbf{Yelp}} & \textit{\textbf{Sci}} & \textit{\textbf{Avg.}}
& \textit{\textbf{CMV}} & \textit{\textbf{ELI5}} & \textit{\textbf{WP}} & \textit{\textbf{TLDR}} & \textit{\textbf{ROCT}} & \textit{\textbf{Avg.}} \\
\midrule
\rowcolor{rowblue}
\multicolumn{14}{c}{\textbf{\textit{Metric-based methods}}} \\
\midrule
\multirow{2}{*}{Entropy} 
& Acc 
& 51.23 & 80.38 & 67.51 & 69.19 & 73.28 & 68.32 
& 65.54 & 70.51 & 72.31 & 67.73 & 63.31 & 67.88 \\
& F1  
& 50.76 & 79.14 & 63.79 & 64.50 & 69.73 & 65.58 
& 70.43 & 71.21 & 72.47 & 70.35 & 71.57 & 71.21 \\
\midrule

\multirow{2}{*}{Rank} 
& Acc 
& 58.54 & 72.57 & 68.80 & 69.13 & 68.44 & 67.50 
& 77.71 & 71.99 & 73.58 & 61.00 & 60.62 & 68.98 \\
& F1  
& 40.07 & 71.08 & 64.77 & 63.70 & 63.21 & 60.57 
& 77.44 & 69.58 & 71.34 & 59.85 & 66.37 & 68.92 \\
\midrule

\multirow{2}{*}{Log-Likelihood} 
& Acc 
& 65.22 & 83.24 & 77.98 & 80.30 & 81.87 & 77.72 
& 86.46 & 82.26 & 82.75 & 71.85 & 62.21 & 77.11 \\
& F1  
& 54.38 & 81.84 & 78.63 & 74.63 & 80.34 & 73.96 
& 85.49 & 79.91 & 80.70 & 74.61 & 72.17 & 78.58 \\
\midrule


\multirow{2}{*}{LRR} 
& Acc 
& 64.84 & 76.95 & 75.00 & 73.88 & 77.34 & 73.60 
& 81.87 & 78.37 & 78.87 & 72.81 & 65.96 & 75.58 \\
& F1  
& 53.11 & 76.83 & 72.49 & 68.76 & 75.51 & 69.34 
& 81.13 & 76.07 & 76.23 & 69.84 & 71.55 & 74.96 \\
\midrule

\multirow{2}{*}{\begin{tabular}[c]{@{}l@{}}Fast-\\DetectGPT\end{tabular}}
& Acc 
& 51.17 & 78.79 & 75.65 & 71.76 & 73.91 & 70.26 
& 73.27 & 75.56 & 62.62 & 71.44 & \textbf{66.56} & 69.89 \\
& F1  
& 54.01  & 77.43  & 76.27  & 71.24  & 74.77  & 70.74 
& 75.00  & 73.60  & 62.15  & 71.73  & 70.40  & 70.58 \\
\midrule
\multirow{2}{*}{Binoculars} 
& Acc 
& 71.77 & 72.06 & 79.65 & 76.89 & 83.75 & 76.82 
& 86.92 & 83.99 & 82.83 & 72.31 & 60.50 & 77.31 \\
& F1  
& 66.77  & 73.84  & 79.29  & 76.22  & 82.43  & 75.71 
& 86.61  & 83.18  & 81.68  & 73.25  & 68.17  & 78.58 \\
\midrule
\rowcolor{rowblue}
\multicolumn{14}{c}{\textbf{\textit{Model-based methods}}} \\
\midrule

\multirow{2}{*}{Roberta$^\dag$} 
& Acc 
& 88.77$_{\text{0.64}}$ & 95.24$_{\text{1.07}}$ & 92.40$_{\text{0.66}}$ & 89.31$_{\text{3.09}}$ & 91.53$_{\text{1.78}}$ & 91.45
& 79.32$_{\text{3.96}}$ & 83.42$_{\text{1.80}}$ & 80.64$_{\text{3.39}}$ & 69.05$_{\text{1.89}}$ & 61.08$_{\text{4.03}}$ & 74.70 \\
& F1  
& 89.61$_{\text{0.48}}$ & 95.35$_{\text{1.18}}$ & 92.74$_{\text{0.54}}$ & 90.05$_{\text{2.46}}$ & 91.93$_{\text{1.48}}$ & 91.94 
& 83.12$_{\text{2.70}}$ & 85.45$_{\text{1.35}}$ & 83.68$_{\text{2.39}}$ & 75.88$_{\text{1.10}}$ & 71.86$_{\text{2.10}}$ & 79.99 \\


\midrule
\multirow{2}{*}{Ghostbuster} 
& Acc 
& 82.44$_{\text{7.94}}$ & 90.79$_{\text{5.58}}$ & 89.99$_{\text{3.98}}$ & 89.67$_{\text{1.91}}$ & 91.77$_{\text{1.62}}$ & 88.93 
& 74.13$_{\text{3.46}}$ & 80.58$_{\text{1.05}}$ & 80.28$_{\text{4.50}}$ & 65.06$_{\text{5.08}}$ & 60.32$_{\text{5.28}}$ & 72.07 \\
& F1  
& 81.84$_{\text{8.89}}$ & 90.79$_{\text{5.57}}$ & 89.89$_{\text{4.14}}$ & 89.63$_{\text{1.97}}$ & 91.76$_{\text{1.63}}$ & 88.78 
& 73.25$_{\text{2.85}}$ & 80.17$_{\text{1.26}}$ & 79.53$_{\text{5.06}}$ & 60.84$_{\text{7.63}}$ & 52.96$_{\text{9.04}}$ & 69.35 \\
\midrule
\multirow{2}{*}{PeCoLA$^\dag$} 
& Acc 
& 88.67$_{\text{1.74}}$ & 96.04$_{\text{0.41}}$ & 91.67$_{\text{0.97}}$ & 91.81$_{\text{0.67}}$ & 92.21$_{\text{0.85}}$ & 92.08 
& 88.83$_{\text{4.18}}$ & 88.32$_{\text{3.06}}$ & 87.87$_{\text{3.61}}$ & 67.73$_{\text{1.06}}$ & 61.51$_{\text{1.22}}$ & 78.85 \\
& F1  
& 88.57$_{\text{1.81}}$ & 96.02$_{\text{0.45}}$ & 91.65$_{\text{0.98}}$ & 91.80$_{\text{0.67}}$ & 92.48$_{\text{0.95}}$ & 92.10
& 88.68$_{\text{4.36}}$ & 88.21$_{\text{3.21}}$ & 87.69$_{\text{3.80}}$ & 65.28$_{\text{2.26}}$ & 55.30$_{\text{1.96}}$ & 77.03 \\
\midrule
\multirow{2}{*}{EAGLE$^\dag$} 
& Acc 
& 88.08$_{\text{3.22}}$ & 95.80$_{\text{0.62}}$ & 92.08$_{\text{1.15}}$ & 91.36$_{\text{1.40}}$ & 90.47$_{\text{4.10}}$ & 91.56 
& 78.54$_{\text{1.78}}$ & 82.31$_{\text{4.58}}$ & 82.84$_{\text{4.43}}$ & 65.85$_{\text{3.10}}$ & 57.36$_{\text{3.01}}$ &  73.38\\
& F1  
& 89.25$_{\text{2.61}}$ & 95.69$_{\text{0.66}}$  & 92.39$_{\text{1.03}}$ & 91.64$_{\text{1.12}}$ & 91.17$_{\text{3.27}}$  &  92.03
& 82.13$_{\text{1.24}}$  & 84.70$_{\text{3.24}}$  & 85.30$_{\text{3.26}}$  & 74.13$_{\text{1.74}}$ & 69.97$_{\text{1.41}}$  & 79.25 \\
\midrule
\multirow{2}{*}{ImBD} 
& Acc 
& 88.77$_{\text{0.95}}$ & 50.0$_{\text{0.20}}$ & 74.87$_{\text{1.72}}$ & 74.85$_{\text{1.46}}$ & 89.71$_{\text{0.47}}$ & 75.64 
& 88.65$_{\text{1.54}}$ & 77.35$_{\text{1.52}}$ & 84.37$_{\text{3.49}}$ & 54.66$_{\text{0.83}}$ & 50.02$_{\text{0.20}}$ & 71.01 \\
& F1  
& 88.85$_{\text{1.22}}$ & 66.67$_{\text{1.56}}$ & 79.65$_{\text{1.06}}$ & 79.82$_{\text{0.88}}$ & 89.06$_{\text{0.59}}$ & 80.81 
& 89.78$_{\text{1.25}}$ & 81.49$_{\text{1.00}}$ & 86.52$_{\text{2.66}}$ & 68.70$_{\text{0.37}}$ & 66.66$_{\text{0.32}}$ & 78.63 \\
\midrule
\multirow{2}{*}{MoSEs} 
& Acc 
& 86.39$_{\text{0.63}}$ & 87.87$_{\text{0.54}}$ & 86.30$_{\text{0.39}}$ & 82.09$_{\text{1.58}}$ & 91.17$_{\text{2.23}}$ & 86.76 
& 86.90$_{\text{1.78}}$ & 79.99$_{\text{2.90}}$ & 83.06$_{\text{2.17}}$ & 71.63$_{\text{1.86}}$ & 61.55$_{\text{2.24}}$ & 76.63 \\
& F1  
& 86.14$_{\text{0.64}}$ & 88.60$_{\text{0.58}}$  & 86.16$_{\text{0.47}}$ & 82.96$_{\text{1.89}}$ & 91.22$_{\text{2.64}}$  & 87.02 
& 86.61$_{\text{1.64}}$  & 78.98$_{\text{2.64}}$  & 82.14$_{\text{2.02}}$  & 72.12$_{\text{1.61}}$ & 48.35$_{\text{2.44}}$  & 73.64 \\
\midrule
\rowcolor{rowblue}
\multicolumn{14}{c}{\textbf{\textit{Domain generalization methods}}} \\
\midrule
\multirow{2}{*}{MSCL$^\dag$} 
& Acc 
& 87.50$_{\text{1.91}}$ & 95.87$_{\text{0.99}}$ & 91.60$_{\text{1.85}}$ & 90.33$_{\text{0.95}}$ & 92.71$_{\text{1.00}}$ & 91.60 
& 81.56$_{\text{5.8}}$ & 83.51$_{\text{3.08}}$ & 82.46$_{\text{4.21}}$ & 69.13$_{\text{3.72}}$ & 61.53$_{\text{2.39}}$ & 75.64 \\
& F1  
& 88.12$_{\text{1.78}}$ & 95.76$_{\text{1.04}}$ & 91.86$_{\text{1.59}}$ & 90.39$_{\text{0.72}}$ & 92.74$_{\text{0.99}}$ & 91.77
& 84.30$_{\text{4.29}}$ & 85.35$_{\text{2.17}}$ & 84.58$_{\text{3.09}}$ & 75.74$_{\text{2.17}}$ & 71.94$_{\text{1.16}}$ & 80.38 \\
\midrule
\multirow{2}{*}{TACIT} 
& Acc 
& 85.23$_{\text{0.78}}$ & 94.13$_{\text{0.77}}$ & 86.40$_{\text{0.87}}$ & 87.89$_{\text{0.67}}$ & 88.36$_{\text{0.96}}$ & 88.40 
& 84.88$_{\text{2.52}}$ & 81.09$_{\text{1.55}}$ & 87.22$_{\text{1.78}}$ & 64.73$_{\text{1.63}}$ & 57.99$_{\text{2.60}}$ & 75.18 \\
& F1  
& 85.76$_{\text{0.57}}$ & 94.18$_{\text{0.72}}$ & 87.09$_{\text{0.59}}$ & 88.34$_{\text{0.50}}$ & 88.79$_{\text{0.78}}$ & 88.83 
& 86.62$_{\text{1.97}}$ & 83.41$_{\text{1.05}}$ & 88.22$_{\text{1.38}}$ & 73.08$_{\text{0.82}}$ & 70.04$_{\text{1.23}}$ & 80.27 \\
\midrule
\multirow{2}{*}{MGT-Prism$^\dag$} 
& Acc 
& 88.76$_{\text{2.63}}$ & 96.28$_{\text{1.13}}$ & 91.81$_{\text{1.90}}$ & 91.14$_{\text{1.34}}$ & 91.93$_{\text{2.06}}$ & 91.98
& 89.08$_{\text{1.94}}$ & 87.34$_{\text{3.64}}$ & 88.58$_{\text{2.95}}$ & 70.53$_{\text{1.32}}$ & 65.47$_{\text{2.22}}$ &  80.20 \\
& F1  
& 89.02$_{\text{2.50}}$ & 96.23$_{\text{1.11}}$ & 92.57$_{\text{0.81}}$ & 91.16$_{\text{1.27}}$ & 92.06$_{\text{1.86}}$ & 92.21 
& 88.14$_{\text{3.59}}$ & 88.39$_{\text{2.85}}$ & 89.48$_{\text{2.34}}$ & 74.34$_{\text{0.95}}$ & \textbf{73.56$_{\text{1.20}}$} & 82.78 \\
\midrule
\multirow{2}{*}{\modelname{}$^\dag$} 
& Acc 
& \textbf{91.70$_{\text{0.74}}$} & \textbf{96.30$_{\text{0.28}}$} & \textbf{93.15$_{\text{1.12}}$} & \textbf{93.09$_{\text{0.52}}$} & \textbf{92.90$_{\text{0.67}}$} & \textbf{93.43} 
& \textbf{92.13$_{\text{2.28}}$} & \textbf{91.79$_{\text{0.95}}$} & \textbf{93.35$_{\text{1.42}}$} & \textbf{72.63$_{\text{1.06}}$} & 62.43$_{\text{0.98}}$ & \textbf{82.46} \\
& F1  
& \textbf{91.67$_{\text{0.93}}$} & \textbf{96.35$_{\text{0.54}}$} & \textbf{93.20$_{\text{0.95}}$} & \textbf{92.94$_{\text{0.65}}$} & \textbf{93.28$_{\text{0.53}}$} & \textbf{93.49} 
& \textbf{92.63$_{\text{1.88}}$} & \textbf{92.02$_{\text{0.79}}$} & \textbf{94.39$_{\text{1.67}}$} & \textbf{76.89$_{\text{1.06}}$} & 72.59$_{\text{0.10}}$ & \textbf{85.70} \\
\bottomrule
\end{tabular}
}
\caption{Comparison of \modelname{} with a broader set of baselines on MGT detection across 10 test domains, using Accuracy and F1 (\%) as evaluation metrics.}
\label{main_exp_broader}
  \vspace{-10pt}
\end{table*}

\end{document}